\documentclass[journal,transmag]{IEEEtran}

\usepackage{lineno,hyperref}
\usepackage{multirow}
\usepackage{graphicx}
\usepackage{subcaption}
\usepackage{mathtools}
\usepackage[export]{adjustbox}

\usepackage{setspace}

\ifCLASSINFOpdf

\else

\fi

\begin{document}

\title{Masked Face Recognition: Human vs. Machine}

\author{Naser Damer$^{a,b,*}$, Fadi Boutros$^{a,b}$, Marius S\"u\ss{}milch$^{a}$, Meiling Fang$^{a,b}$, \\ Florian Kirchbuchner$^{a}$, Arjan Kuijper$^{a,b}$\\
$^{a}$Fraunhofer Institute for Computer Graphics Research IGD, Darmstadt, Germany\\
$^{b}$Mathematical and Applied Visual Computing, TU Darmstadt,
Darmstadt, Germany\\
Email: naser.damer@igd.fraunhofer.de
}

\maketitle

\begin{abstract}
The recent COVID-19 pandemic has increased the focus on hygienic and contactless identity verification methods. However, the pandemic led to the wide use of face masks, essential to keep the pandemic under control. The effect of wearing a mask on face recognition (FR) in a collaborative environment is a currently sensitive yet understudied issue. Recent reports have tackled this by evaluating the masked probe effect on the performance of automatic FR solutions. However, such solutions can fail in certain processes, leading to performing the verification task by a human expert. This work provides a joint evaluation and in-depth analyses of the face verification performance of human experts in comparison to state-of-the-art automatic FR solutions. This involves an extensive evaluation by human experts and 4 automatic recognition solutions. The study concludes with a set of take-home messages on different aspects of the correlation between the verification behavior of humans and machines.
\end{abstract}


\IEEEpeerreviewmaketitle

\section{Introduction}

In hygiene-sensitive scenarios, such as the current COVID-19 pandemic, the importance of contactless and high-throughput operations is escalating, especially at crowded facilities such as airports.
An existing accurate and contactless identity verification method is FR.
However, covering the face with a facial mask has been forced in public places in many countries during the COVID-19 pandemic to fight the spread of the contagious disease.
This fact can influence the performance, and thus trust, of automatic FR systems and questions their functionality under a situation where the individuals' faces are masked.
More critically, when operations based on automatic FR (e.g. automatic border control (ABC) gates) fail, identity verification becomes the task of human experts (e.g. border guard).
This is also the case when such automatic solutions do not exist (e.g. passport not compliant with ABC gates, remote land borders, regular police checks, etc.).
As this identity verification by human experts requires relatively close contact, wearing masks becomes even more essential for pandemic control.
However, as the masks affect the verification performance of automatic systems, they might also affect the human expert performance, an issue that is not studied yet.

Recent studies have looked into the effect of wearing a mask on automatic FR performance. 
The National Institute of Standards and Technology (NIST) has evaluated the effect of face masks (synthetically created masks) on the performance of several FR systems provided by vendors \cite{ngan2020ongoing}. 
The Department of Homeland Security has conducted a similar evaluation, however on more realistic data \cite{DHS-Masks-2020}. 
A study by Damer et al. \cite{DBLP:conf/biosig/DamerGCBKK20} evaluated the verification performance drop in 3 FR systems when verifying unmasked-to-masked faces, in comparison to verifying unmasked-faces to each other, all with real masks and in a collaborative environment. This study has been later extended, \cite{IETMask} with an extended database and evaluation on both synthetic and real masks. All these studies have concluded with the significant effect of probes wearing a face mask on the recognition performance. However, they did not conduct a study on the human verification performance or the relation of this performance to the automatic recognition performance.

This work provides an in-depth joint evaluation and analysis of the effect of masked-faces on the verification performance of human experts and automatic FR solutions. This comes as a needed effort to enable the development of solutions addressing accurate face verification under these scenarios and influence the design of identity-related processes. To achieve that, this paper presents the following contributions:
\begin{itemize}
    \item A detailed joint evaluation of the face verification performance of human experts in comparison to state-of-the-art automatic FR solutions. This included extended experiments with 12 human experts and the evaluation of 4 automatic FR systems, including one of the top-performing commercial off-the-shelf (COTS) systems.
    \item Uncover the behavior of the human experts and automatic solutions when dealing with the different settings including facial masks by investigating the changes in the produced similarity scores.
    \item Asses the experimental advantages of having a masked reference for the cases where the probe is masked. This is conducted using real and simulated masks, for the human experts in comparison to the automatic recognition systems.
    \item We provide an experimental opinion on the use of synthetically created masks as an alternative for real masks in evaluating the face verification performance of human experts in comparison to automatic solutions. This is conducted under different scenarios, where both or one of the compared faces are masked.
    \item We study the effect of wider and tighter face crops (i.e. the visibility of the face bordering area) on the human expert verification performance on the different experimental evaluations, especially when masked-faces are involved.
    \item We list a condensed and consistent set of observations made by the human experts on the different configurations of the masked-face verification tasks.
\end{itemize}

Our conclusions are summarized in a list of take-home messages in Section \ref{ssec:thm} and they include pointing out, among other observations, the negative effect of masked-faces on the human expert verification performance and the strong correlation of this effect with the automatic solutions. In the rest of the paper, we shortly introduce related works from the literature. We present our database then discuss both, the human expert verification setup and the used automatic FR solutions. We follow that with the experimental setup and a detailed presentation of the results in the form of responses to research questions. We end our results with a list of human experts' observations and a condensed set of take-home messages.

\section{Related works}

Several operational challenges encounter the deployment of FR solutions.
Issues related to attacks on FR systems are considered the most important of these challenges and thus receives most of the research attention. Such attacks can be morphing attacks \cite{DBLP:conf/icb/DamerSZWTKK19}, presentation attacks (spoofing) \cite{DBLP:conf/bmvc/DamerD16}, or different unconventional attacks \cite{DBLP:conf/btas/DamerWBBT0K18}.
However, FR deployability is also affected by factors related to the biometric sample capture and presentation, including face occlusions.
Occluded faces detection is a widely-studied challenge in the domain of computer vision \cite{DBLP:conf/eccv/OpitzWPPB16,DBLP:conf/cvpr/GeLYL17}.
A study by Optiz et al. \cite{DBLP:conf/eccv/OpitzWPPB16} is a clear example of that, as they targeted the accurate detection of occluded faces by presenting a solution based on a novel grid loss.
Ge et al. \cite{DBLP:conf/cvpr/GeLYL17} focused on detecting faces (not FR) with mask occlusions in in-the-wild scenarios.
Their research included face-covering objects in a wider perspective rather than facial masks worn for medical or hygiene reasons.
Such studies are highly relevant to FR because the detection of faces (while wearing masks) is an essential pre-processing step where FR systems might fail, as will be shown later by our experiments.

As discussed, the detection of occluded faces is one of the challenges facing the deployment of face biometric systems.
However, more importantly, the biometric recognition of occluded faces is a more demanding challenge \cite{DBLP:journals/pami/KimCYT05,OU20141559,DBLP:conf/iccv/SongGLLL19}.
One of the recent works to address this issue is the research of Song et al. \cite{DBLP:conf/iccv/SongGLLL19} that aimed at improving the performance of FR under general occlusions.
The approach presented by Song et al. \cite{DBLP:conf/iccv/SongGLLL19} localizing and abandon corrupted feature elements, that might be associated with occlusions, from the recognition process.
A similar approach was later presented in \cite{9495272}, where the separate step of detecting blocked areas in  \cite{DBLP:conf/iccv/SongGLLL19} was integrated within an end-to-end solution.
A very recent work presented by Wang et al. \cite{wang2020masked}  focused on masked-faces. Their work introduced, in a short and under-detailed presentation, crawled databases for FR, detection, and simulated masked-faces.
The authors claim to improve the verification accuracy from 50\% to 95\%, however, did not provide any information on the used baseline, the proposed algorithmic details, or clearly point to the used evaluation database. 
A recent pre-print by Anwar and Raychowdhury \cite{anwar2020masked} has presented a database that includes 296 face images, partially with real masks, of 53 identities. The images in the database can be considered to be captured under in-the-wild conditions, as they are crawled from the internet, and thus do not represent a collaborative FR scenario. The proposed fine-tuning an existing FR network to achieve better evaluation performance.
Very recent works addressed enhancing masked FR accuracy exclusively. Li et al. \cite{DBLP:journals/apin/LiGLL21} proposed to cropping approach that will focus on the upper part of the face. Boutros et al. \cite{DBLP:journals/corr/abs-2103-01716} proposed an approach that will transfer the template of a masked-face into an unmasked-like template through an on-the-top network trained with a especially proposed Self-restrained Triplet Loss. Follow up solutions trained FR models in a way that would promote producing similar face templates for masked an unmasked faces \cite{DBLP:conf/fgr/HuberBKD21,DBLP:conf/fgr/NetoBPDSC21,DBLP:conf/biosig/ErakinDE21,DBLP:conf/biosig/NetoBPSDS021}. This interest in enhancing FR performance on masked faces led to two competitions that attracted a diverse set of academic and industrial participants \cite{DBLP:conf/icb/BoutrosDKRKRKFZ21,DBLP:conf/iccvw/DengGAZZ21}. Further studies by Fang et al. \cite{DBLP:journals/corr/abs-2103-01546} has unraveled the vulnerabilities of face presentation attack detection solutions to presentation attacks of masked-faces or attacks with real masks placed on them and proposed possible technical solutions \cite{DBLP:conf/fgr/FangBKD21}.

On a larger scale, the National Institute of Standards and Technology (NIST), as a part of the ongoing FR Vendor Test (FRVT), has published a specific study (FRVT -Part 6A) on the effect of face masks on the performance on FR systems provided by vendors \cite{ngan2020ongoing}. The NIST study concluded that the algorithm accuracy with masked-faces declined substantially. One of the main study limitations is the use of simulated masked images under the questioned assumption that their effect represents that of real face masks.
The Department of Homeland Security has conducted a similar evaluation, however on more realistic data \cite{DHS-Masks-2020}. They also concluded with the significant negative effect of wearing masks on the accuracy of automatic FR solutions.
A study by Damer et al. \cite{DBLP:conf/biosig/DamerGCBKK20} evaluated the verification performance drop in 3 face biometric systems when verifying unmasked-to-masked faces, in comparison to verifying unmasked-faces to each other. The authors presented limited data (24 subjects), however, with real masks and multiple capture sessions. They concluded by noting the bigger effect of masks on genuine pairs decisions, in comparison to imposter pairs decisions. This study has been extended \cite{IETMask} with a larger database and evaluation on both synthetic and real masks, with further analyses linking the drop in FR performance to the drop in face image quality \cite{DBLP:journals/corr/abs-2112-06592,DBLP:conf/fgr/FuKD21}

Regarding human performance in FR, an early study by O'Toole et al. \cite{Toole06predictinghuman} studied the effect of stucture constraints and viewing parameters on human. The stucture factors included issues like face typicality, gender, and ethnicity, while the viewing parameters included variations in illumination, viewpoint, and motion effect. Later on, a comparison between human and automatic FR was presented in \cite{PHILLIPS201474}. The authors concluded with the challenges that automatic FR faces to reach human performance. However, the advances in deep learning approaches have later placed automatic FR performance ahead of human performance \cite{DBLP:conf/aaai/LuT15}. Other interesting and well-studied aspects of human performance in FR addressed the issue of other-race effect \cite{structural_other_race,10.1145/1870076.1870082}, where human recognizers showed better performance on faces of familiar demographic groups. Moreover, a recent study has looked into the outcome of automatic algorithm cognitive effect on the human decision-making bias \cite{Howard2020HumanalgorithmTI}.

However, all these studies did not explicitly evaluate the degradation in the verification performance of human experts when comparing occluded faces, in particular faces with hygienic face masks (masked-to-unmasked-faces or masked-to-masked-faces), which is essential to many security processes and is one of the main focus points of this work.

Under the current COVID-19 pandemic, an explicitly gathered database and the evaluation of real face masks on collaborative face verification by human experts and its correlation to automatic systems is essential. This research gap, as cleared above, also includes the need of comparing the appropriateness of using simulated face masks for face verification performance and assessing the performance drop when comparing masked-face pairs by both, software solutions and human experts. These issues are all addressed in this work.

\section{The database}
\label{sec:data}

The collected database aims at enabling the analyses of FR performance on masked-faces and motivates future research in the domain of masked FR.
Our data aims to represent a collaborative, however varying, scenario.
The targeted scenario is that of unlocking personal devices or identity verification at border control. 
This is essential as human experts regularly verify the identity of travelers, in comparison to their passport images, at border checkpoints.
This motivated the variations in masks, illumination, and background in the presented database.

The participants were requested to capture the data on three different days, not necessarily consecutive.
Each of these days is considered one session.
In each of these sessions/days, three videos, with a minimum duration of 5 seconds, are collected by the subjects.
The videos are collected from static webcams (not handheld), while the users are requested to look at the camera, simulating a login scenario.
The data is collected by participants at their residences during the pandemic-induced home-office period.
The data is collected during the day (indoor, day-light) and the participants were requested to remove eyeglasses when considered to have very thick frames.
To simulate a realistic scenario, no restrictions on mask types or background, nor any other restrictions, were imposed.
The three captured videos each day were as given:
1) not wearing a mask and with no additional electric lighting (illumination), this scenario is referred to as baseline (BL). 2) wearing a mask and with no additional electric lighting (illumination), this scenario is referred to as baseline (M1). 3) wearing a mask with the existing electric lighting (illumination) in the room turned on, this scenario is referred to as baseline (M2).
Given that wearing a mask might result in varying shadow and reflection patterns, the M2 scenario is considered to investigate the unknown effect of illumination variation in masked FR. All the subject provided their informed consent to use the data for research purposes.

\begin{table*}[ht!]
\scriptsize
\centering
\begin{tabular}{|l|l|l|l|l|l|l|l|l|l|}
\hline
Session            & \multicolumn{4}{l|}{Session 1: References} & \multicolumn{5}{l|}{Session 2 and 3:   Probes} \\ \hline
Data split         & BLR          & M1R          & M2R          & SMR          & BLP       & M1P       & M2P       & M12P       & SMP   \\ \hline
Illumination       & No           & No           & Yes          & No           & No        & No        & Yes       & Both       & No    \\ \hline
Number of Captures & 480          & 480          & 480          & 480          & 960       & 960       & 960       & 1920       & 960   \\ \hline
\end{tabular}
\caption{An overview of the database structure.}
\label{tab:DB}
\end{table*}

\begin{figure*}[ht!]
     \centering
     \begin{subfigure}[b]{0.24\textwidth}
         \centering
         \includegraphics[width=\textwidth]{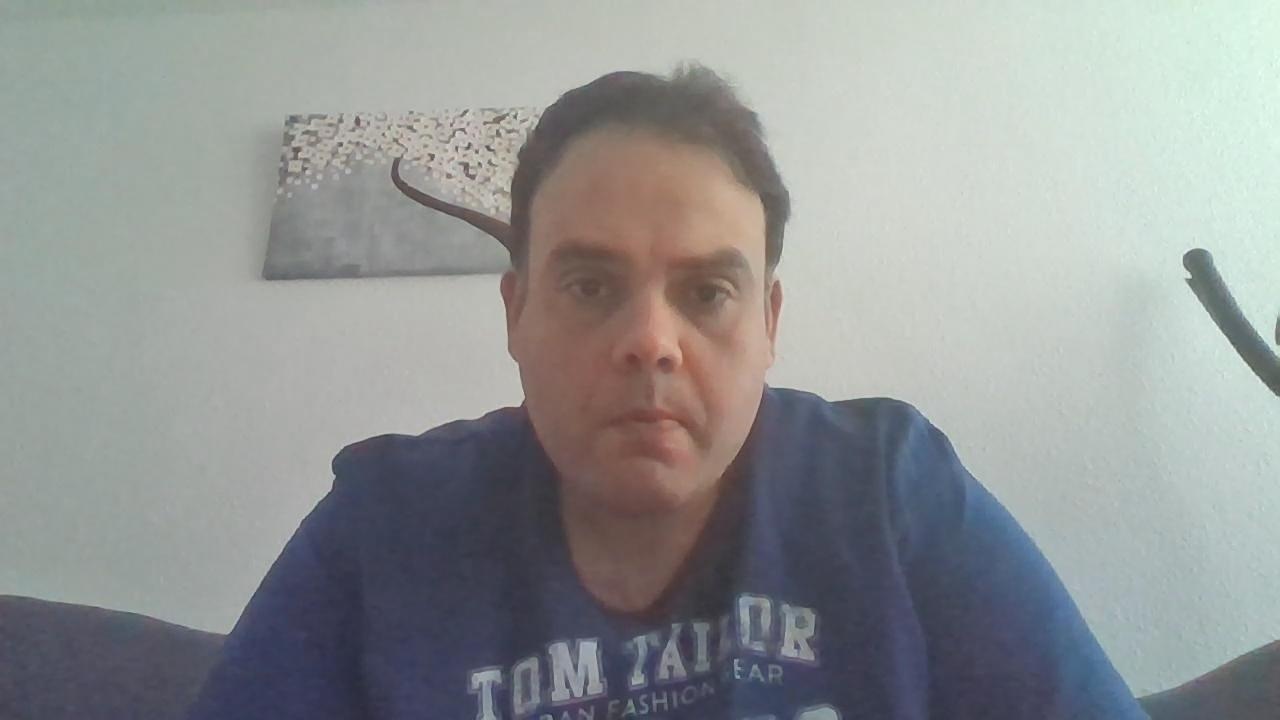}
         \\
         \includegraphics[width=\textwidth]{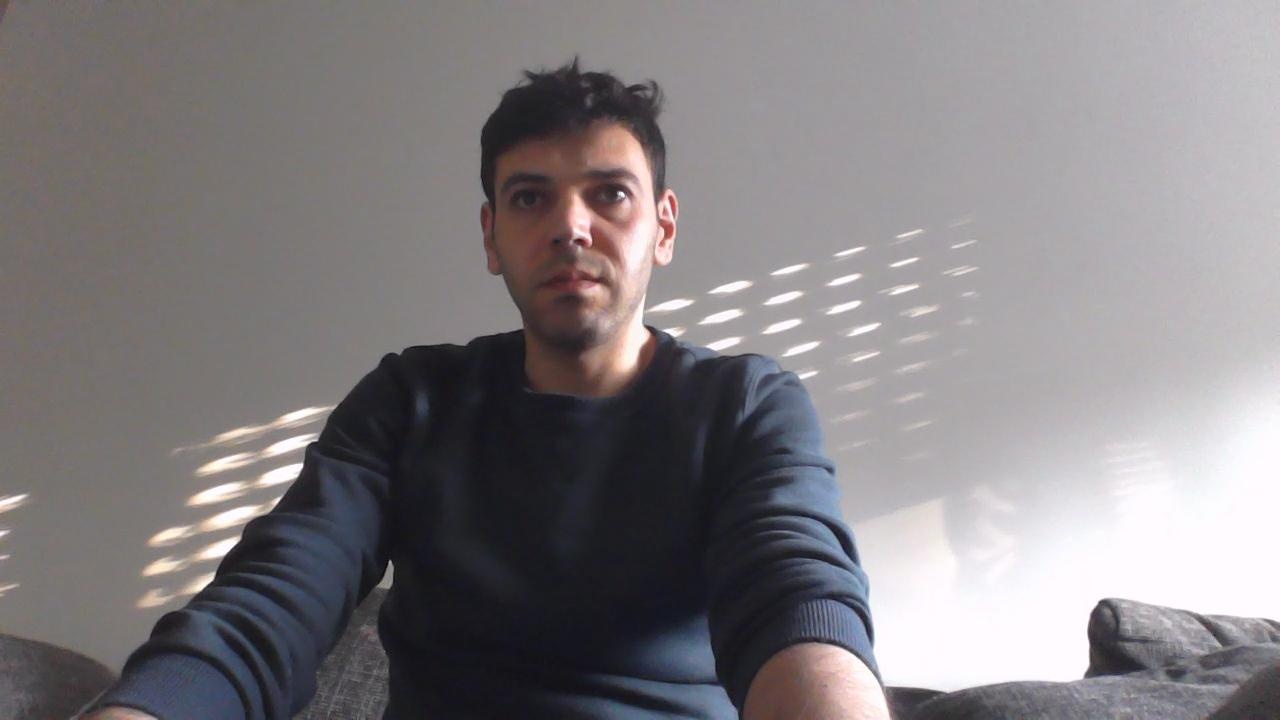}
         \caption{BL}
         \label{fig:samp:BL}
     \end{subfigure}
     \hfill
     \begin{subfigure}[b]{0.24\textwidth}
         \centering
         \includegraphics[width=\textwidth]{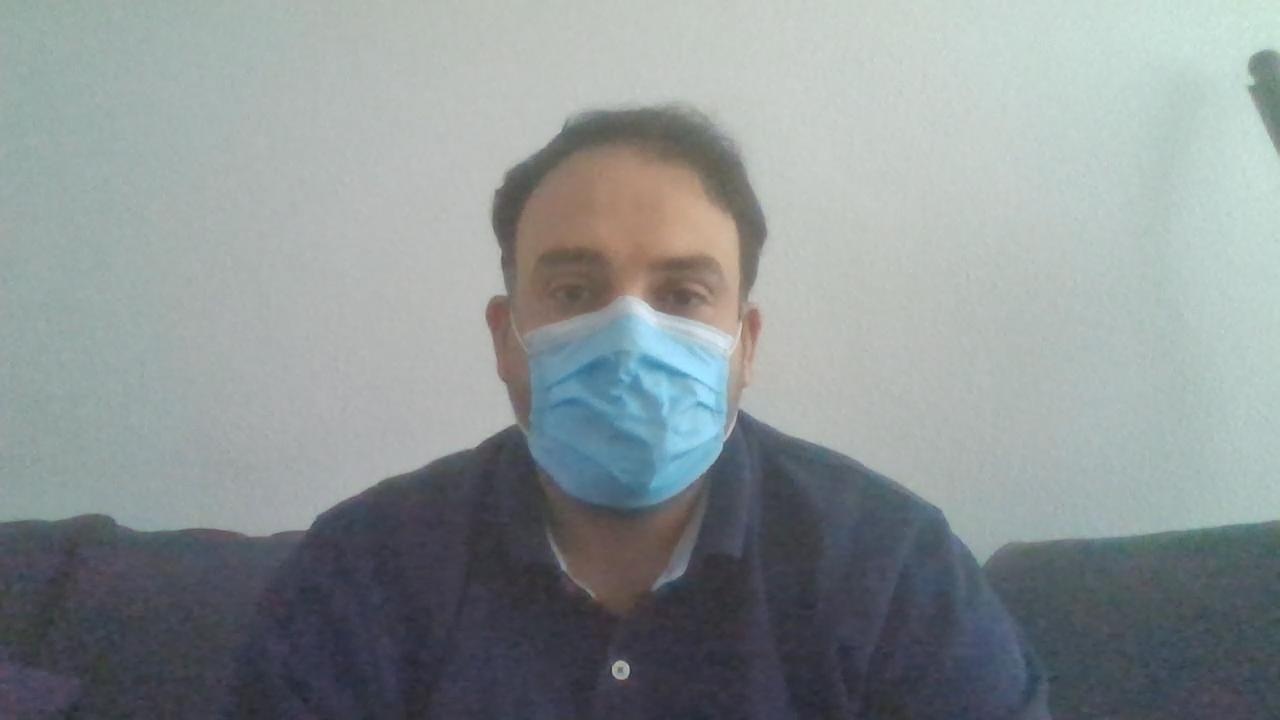}
         \includegraphics[width=\textwidth]{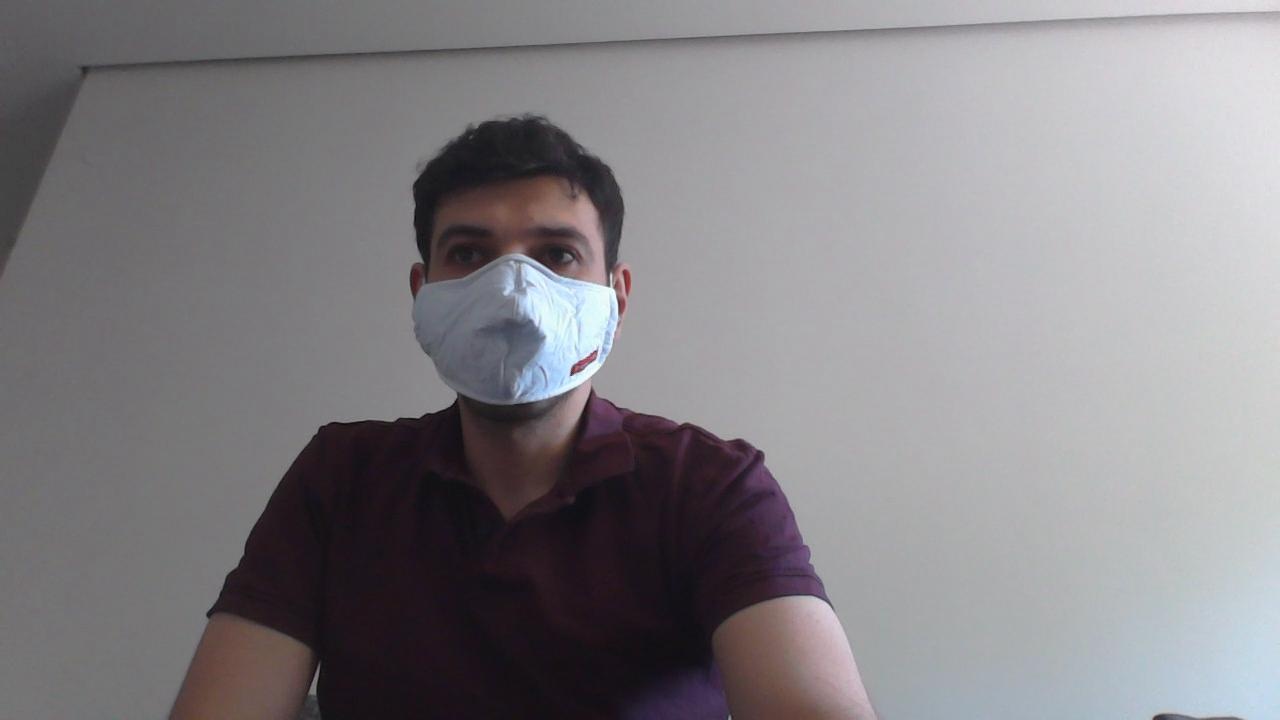}
         \caption{M1}
         \label{fig:sam:M1}
     \end{subfigure}
     \hfill
     \begin{subfigure}[b]{0.24\textwidth}
         \centering
         \includegraphics[width=\textwidth]{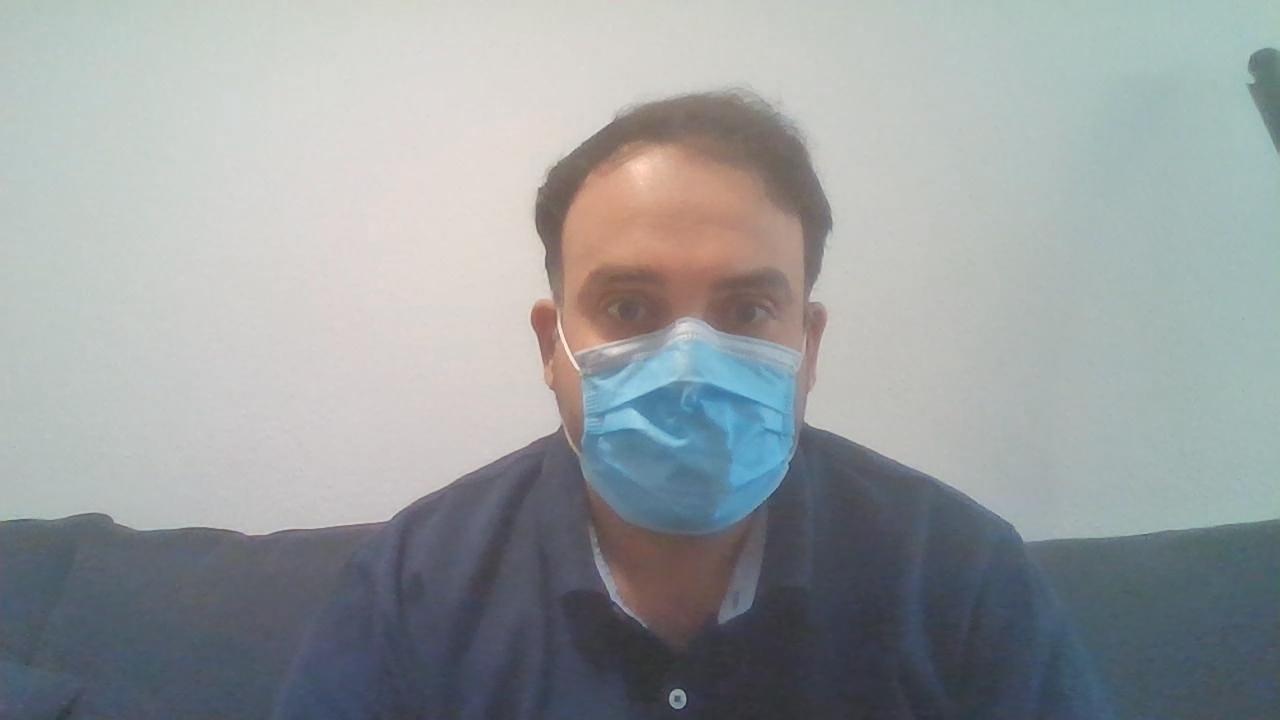}
         \includegraphics[width=\textwidth]{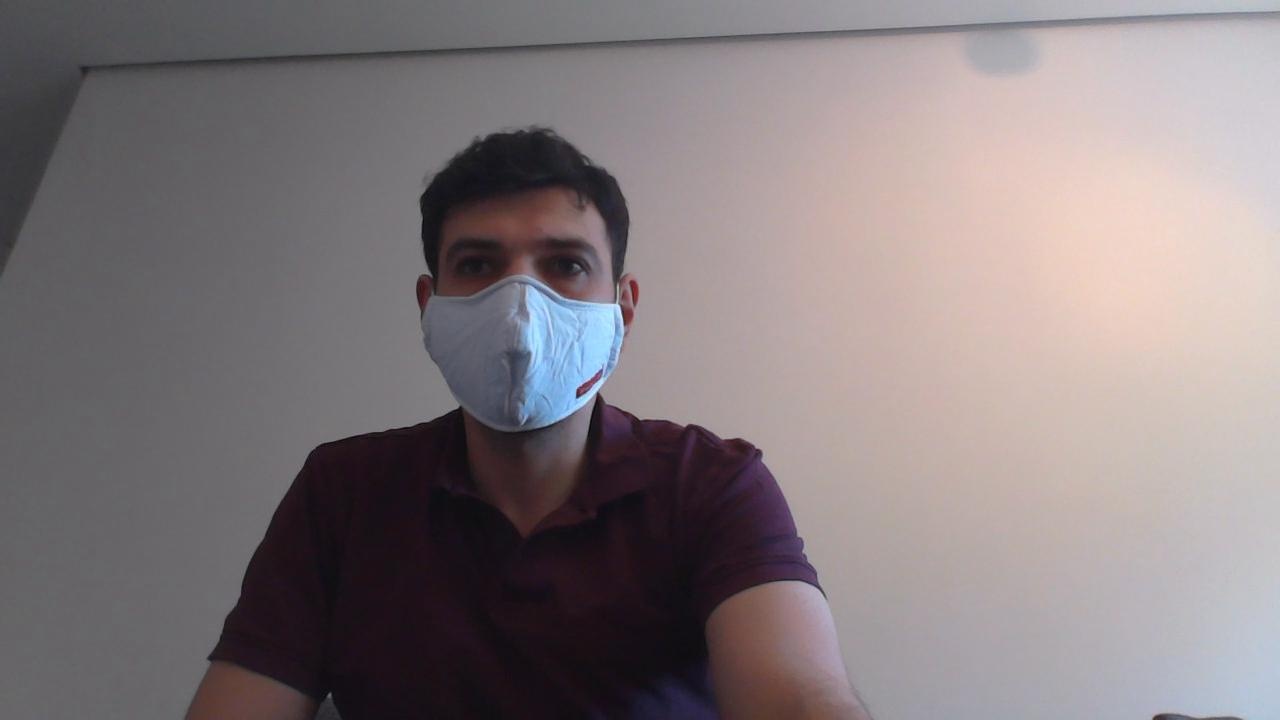}
         \caption{M2}
         \label{fig:sam:M2}
     \end{subfigure}
     \hfill
     \begin{subfigure}[b]{0.24\textwidth}
         \centering
         \includegraphics[width=\textwidth]{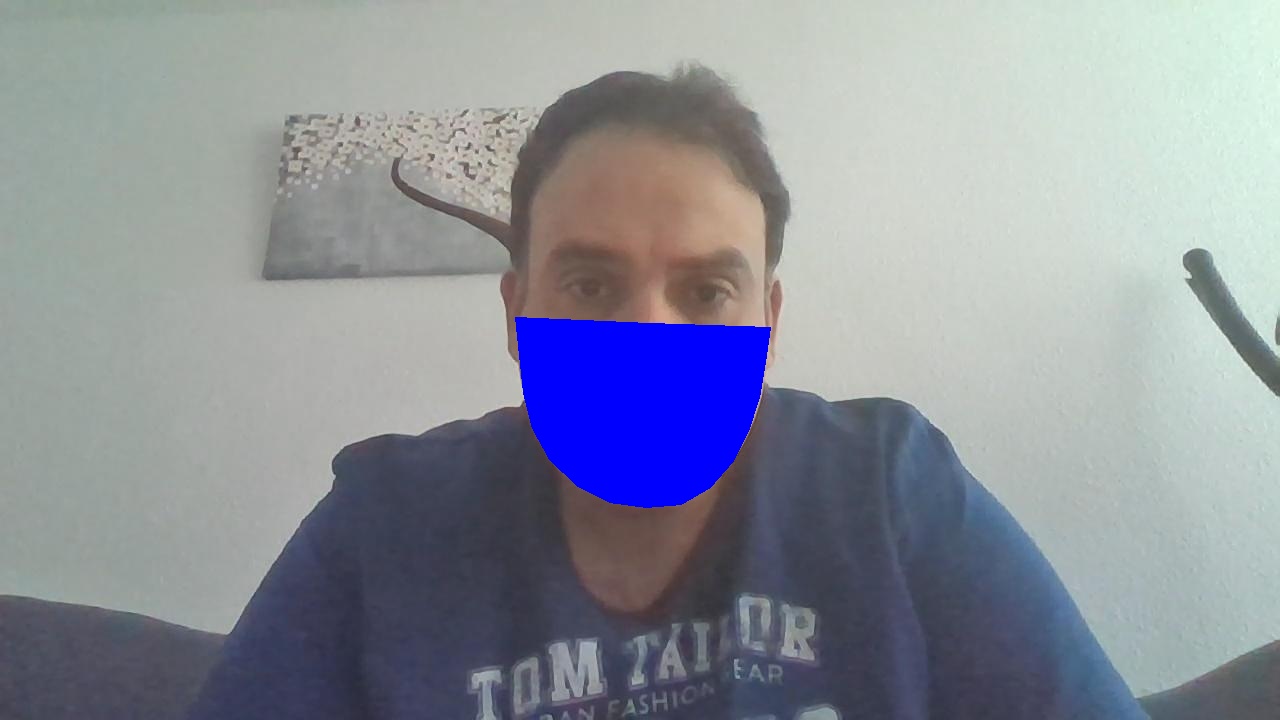}
         \includegraphics[width=\textwidth]{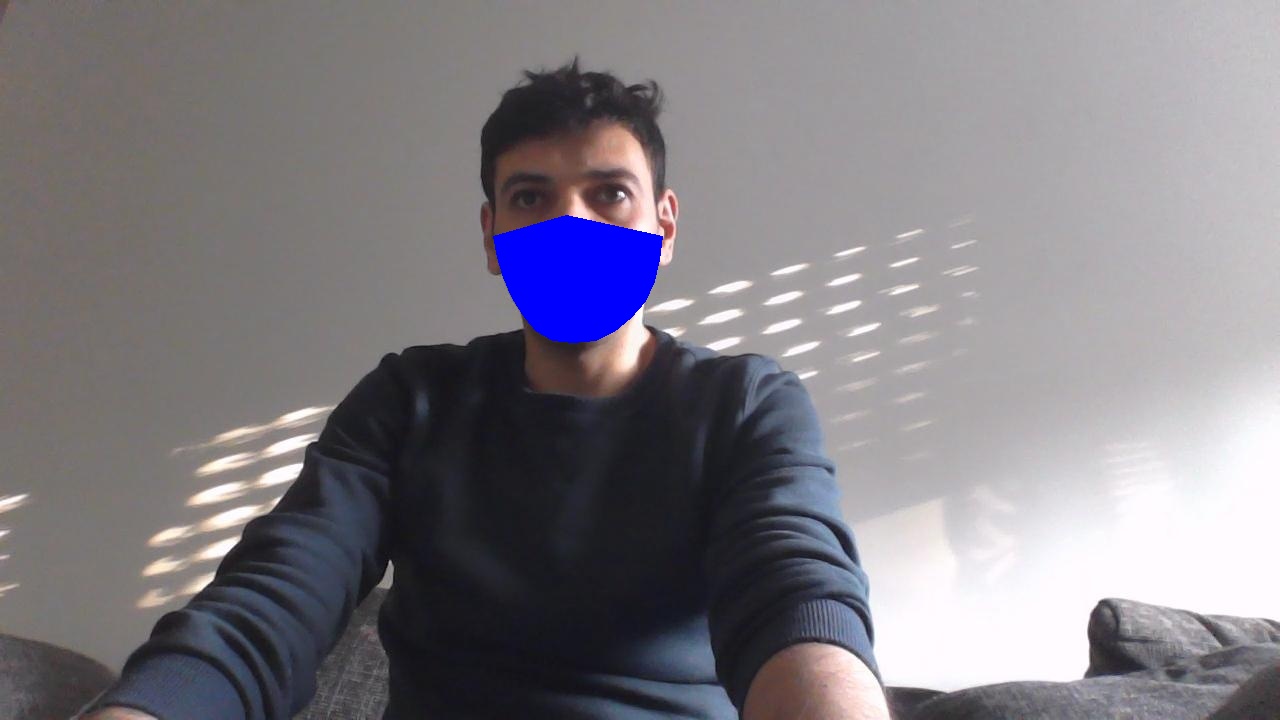}
         \caption{SM}
         \label{fig:sam:SM}
     \end{subfigure}    
        \caption{Samples of the collected database from the three capture types (BL, M1, M2, SMP).}
        \label{fig:samples}
\end{figure*}

The reference (R) data is the data from the first session (day), consisting of the baseline reference (BLR), the mask reference of the first capture scenario (M1R), and the mask reference of the second capture scenario (M2R).
The probe data (P) is the data of the second and third sessions (days) including the no-mask baseline probe (BLP), the mask probe of the first capture scenario (M1P), mask probe of the second capture scenario (M2P), and the combined probe data (M1P and M2P) noted as M12P. 
The first second of each video was neglected to avoid possible biases related to the subject interaction with the capture device.
After the neglected one second, three seconds were considered.
From these three seconds, 10 frames are extracted with a gap of 9 frames between each consecutive frame, knowing that all videos are captured at a frame rate of 30 frames per second.
To enable the comparative evaluation of the synthetically added masks, the data is augmented with additional subsets that include the images with simulated masks. As synthetically masked references, simulated masks are added to the BLR images to create the simulated mask reference subset (SMR). For probes, the BLP subset is augmented with simulated masks as well to create a simulated mask probes subset (SMP). The details of adding the simulated masks are presented in Section \ref{sec:simulated_mask}.   

The total number of participants in the database is 48, in comparison to 24 in the first version of the database \cite{DBLP:conf/biosig/DamerGCBKK20}, the current version of the database was used to evaluate FR performance when faces are masked with real and simulated mask \cite{IETMask}, as well as evaluating novel solutions to enhance the performance of automatic masked FR \cite{DBLP:journals/corr/abs-2103-01716}. All subjects participated in all three required sessions.
Table \ref{tab:DB} presents an overview of the database structure as a result of the number of participants, sessions, and extracted frames from each video.
Figure \ref{fig:samples} shows samples of the database.

\section{Face verification by human experts}

In this section, we present our evaluation of face verification by human experts.
The evaluation process aimed to imitate the process performed by border guards at border checks, i.e. compare a live (probe) image to a reference image (typically on a passport).
A set of 12 human experts were selected, all participants have experience in developing biometric solutions, and thus have a good understanding of the identity verification process.
The human experts were not included in the used database, nor are acquainted with any of the subjects in the database.
This was done to assure a fair verification evaluation without any biases that might be introduced by the prior knowledge of certain faces.

\begin{figure*}[ht!]
     \centering
     \begin{subfigure}[b]{0.20\textwidth}
         \centering
         \includegraphics[width=1.6cm,height=2cm,]{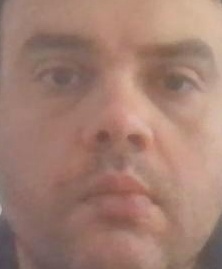}\\
         \includegraphics[width=1.6cm,height=2cm,]{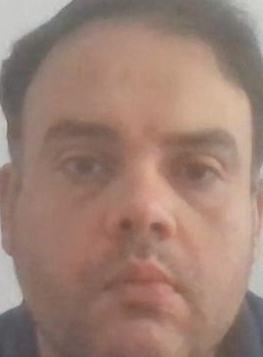}
         \caption{BL}
         \label{fig:cro:BL}
     \end{subfigure}
     \begin{subfigure}[b]{0.2\textwidth}
         \centering
         \includegraphics[width=1.6cm,height=2cm]{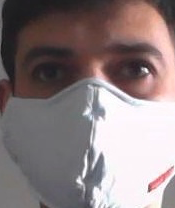}\\
         \includegraphics[width=1.6cm,height=2cm]{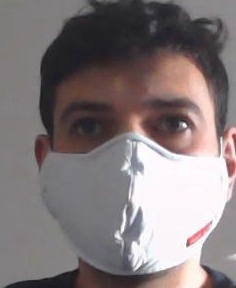}
         \caption{M1}
         \label{fig:cro:M1}
     \end{subfigure}
     \begin{subfigure}[b]{0.2\textwidth}
         \centering
         \includegraphics[width=1.6cm,height=2cm]{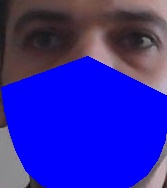}\\
         \includegraphics[width=1.6cm,height=2cm]{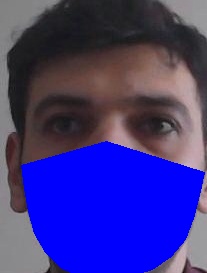}
         \caption{M2}
         \label{fig:cro:M2}
     \end{subfigure}
        \caption{Samples of crop-1 (top row) and crop-2 (bottom row) for unmasked-faces (BL), masked-faces (M1), and synthetically masked-faces (M2). One can notice the larger visible area in crop-2, including the hair style.}
        \label{fig:crops}
\end{figure*}

Our database, presented in Section \ref{sec:data}, was collected over a period of a few weeks. However, the identity verification process in processes such as border control may include longer time gaps between the capture of the reference and probe.
This time gap increases the possibility of larger changes in the hairstyles and even hairlines, which should not affect the expert's verification decision.
To simulate such issues, and evaluate the role of a wider visible face view, we present the verification pairs from our data to the experts in two different crop styles, crop-1 and crop-2.
Both crops are based on an initial detection by the Multi-task Cascaded Convolutional Networks (MTCNN) \cite{zhang2016joint}. For crop-1, the initial MTCNN detection was tightened by removing 10\% of the width from both the right and left borders, and 10\% of the height from both the bottom and the top borders. Crop-2 was performed by restricting the amount of crop of the original detection by removing 5\% of the width from both the right and left borders, and 5\% of the height from both the bottom and the top borders. Examples of both crops are shown in Figure \ref{fig:crops} from different data subsets, crop-1 on the top and crop-2 on the bottom of the figure.

\begin{figure*}[ht!]
     \centering
      \includegraphics[width=0.6\textwidth]{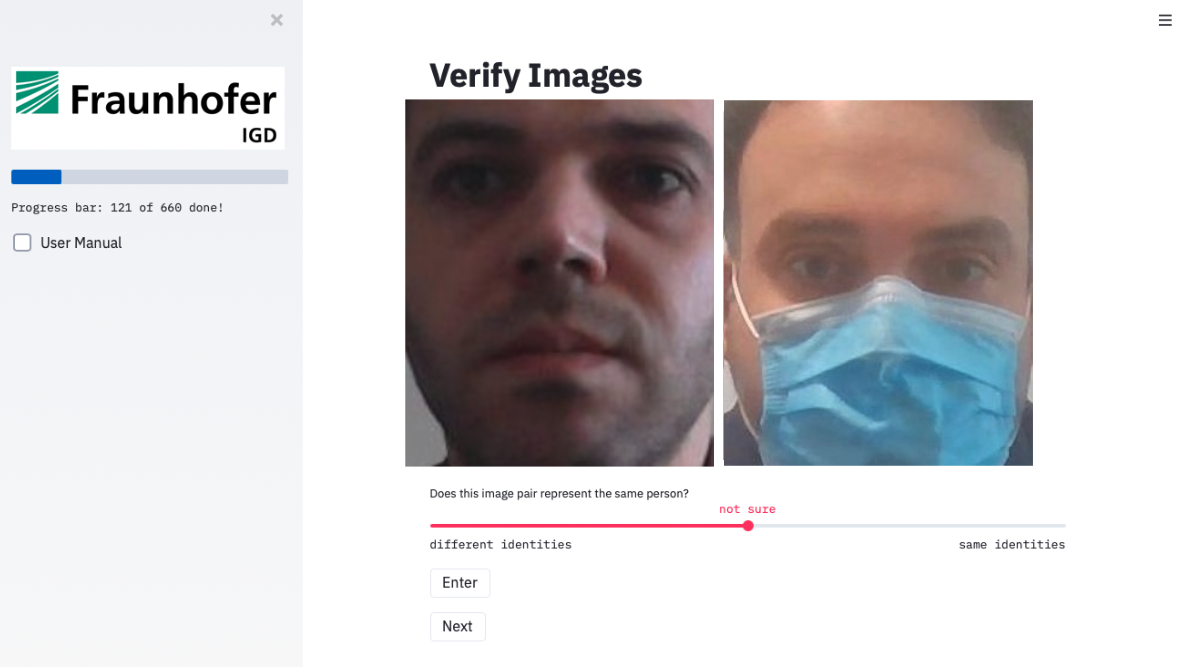}
      \caption{A screenshot of the verification application that the human experts use to visualize the image pairs and enter their similarity decision as an integer number between 0 (full confidence in imposter pair decision) and 10 (full confidence in genuine pair decision), while 5 indicates an inability to make a decision and the other values are different degrees of confidence in an imposter of genuine decision.}
      \label{fig:app}
\end{figure*}

An application was developed to present the verification pairs to the human experts and collect their decisions. In this application, a pair of face images are shown and the human expert is requested to move a slider to a number between "0" and "10" to reflect the identity similarity between the image pair. A decision of "0" indicates full confidence that the two faces belong to different identities, a decision of "10" indicates full confidence that the two faces belong to the same identity, and a decision of "5" indicates a complete inability to make a decision. The other integer numbers can indicate different confidence degrees in the face pair being an imposter (1 to 4) or genuine (6 to 9) pair.
The user confirms the decision by pressing an "Enter" bottom and pressing "next" to the next verification pair.
The order of the presented pairs for verification is selected randomly from all the considered data reference-probe combinations and crop styles. This random ordering is performed to avoid any biases that may be induced by a static ordering of the crop or data combination types. A screenshot of the application is shown in Figure \ref{fig:app}.

The following reference-probe combinations of face pairs were evaluated by human experts: BLR-BLP, BLR-M12P, BLR-SMP, M12R-M12P, SMR-M12P, SMR-SMP. These combinations are the same that are evaluated by automatic FR solutions. For each of these combinations, and for each human expert, 40 genuine pairs and 120 imposter pairs were selected randomly (from all identities and images). More imposter pairs were selected than genuine pairs because the possible variation in the imposter space is larger, i.e. there are more possible imposter comparisons than genuine ones.
12 human experts participated in the evaluation resulting in a total of 480 genuine scores and 1440 imposter scores per reference-probe combination type. Given the 6 combinations, our human expert experiment resulted in a total of 2880 genuine scores and 8640 imposter scores for each crop type. This set of experiments are repeated for each of the cropping styles, crop-1 and crop-2. The verification results from the human experts on crop-1 and crop-2 will be noted as "Human-c1" and "Human-c2", respectively.

\section{Face verification by machine}

We analyze the performance of four FR algorithms to present a wide view of the effect of wearing a mask on FR performance.
These algorithms are the ArcFace \cite{DBLP:conf/cvpr/DengGXZ19}, VGGFace \cite{Parkhi15}, SphereFace \cite{DBLP:conf/cvpr/LiuWYLRS17}, and the COTS solution from Neurotechnology \cite{neurotechnology12}. The FR algorithms choices were motivated by the need to make generalized conclusions on the relation between the verification performance behaviors of FR and human experts when dealing with masked faces. Thus, the choices aimed at analyzing FR algorithms at different levels of maturity/performance along with a market-mature COTS.
The following presents detailed information of the selected algorithms:

\paragraph{SphereFace:}
SphereFace is chosen as it achieved competitive verification accuracy on Labeled Face in the Wild (LFW) \cite{LFWTech} $99.42\%$ and Youtube Faces (YTF) \cite{wolf2011face} $95.0\%$.  SphereFace contains 64-CNN layers trained on the CASIA-WebFace dataset \cite{DBLP:journals/corr/YiLLL14a}. SphereFace is trained with angular Softmax loss function (A-Softmax). The key concept behind  A-Softmax loss is to learn discriminative features from the face image by formulating the Softmax as angular computation between the embedded features vector $X$ and their weights $W$.

\paragraph{ArcFace:}
ArcFace scored state-of-the-art performance on several FR evaluation benchmarks such as LFW $99.83\%$ and YTF $ 98.02\%$. ArcFace introduced Additive Angular Margin loss (ArcFace) to improve the discriminative ability of the FR model. We deployed ArcFace based on ReseNet-100 \cite{DBLP:conf/cvpr/HeZRS16} architecture pretrained on refined version of the MS-Celeb-1M dataset \cite{DBLP:conf/eccv/GuoZHHG16} (MS1MV2).

\paragraph{VGGFace:}
VGGFace is one of the earliest FR models that achieved a competitive verification accuracy on LFW  ($98.95\%$) and YTF ($97.3\%$) face benchmarks using simple network architecture trained on a public database (2.6M of images of 2.6K identities). The network architecture is based on VGG model \cite{DBLP:journals/corr/SimonyanZ14a} trained with Softmax loss and fine-tuned using triplet loss \cite{DBLP:conf/cvpr/SchroffKP15}.

\paragraph{COTS:} We used the MegaMatcher 12.1 SDK \cite{neurotechnology12} from the vendor Neurotechnology. We chose this COTS product as Neurotechnology achieved one of the best performances in the recent NIST report addressing the performance of vendor face verification products \cite{nist2020}. All the face quality thresholds were set to zero to minimize neglecting masked-faces. The full processes of detecting, aligning, feature extraction, and matching are part of the COTS and thus we are not able to provide their algorithmic details. Matching two faces by the COTS produces a similarity score. 

For the three academic systems, the widely-used MTCNN \cite{zhang2016joint} solution is used, as recommended in \cite{DBLP:conf/cvpr/LiuWYLRS17}, to detect (crop) and align (affine transformation) the face. 

ArcFace and SphereFace network process the input aligned and cropped image and produce a feature vector of the size 512 from the last network layer. The VGG produces a feature vector of the size 4096 from the third to last output layer as recommended by the authors \cite{Parkhi15}.
To compare two faces, a distance is calculated between their respective feature vectors. This is calculated as Euclidean distance for ArcFace  and VGGFace features, as recommended in \cite{DBLP:conf/cvpr/DengGXZ19} and \cite{Parkhi15} respectively, and as Cosine distance for SphereFace features, as recommended in \cite{DBLP:conf/cvpr/LiuWYLRS17}. The Euclidean distance (dissimilarity) is complemented to illustrate a similarity score and the Cosine distance shows a similarity score by default. The four systems will be noted in the results as "Arc" for ArcFace, "Sphere" for SphereFace, "VGG" for VGGFace, and "COTS" for the commercial solution.

\section{Experimental setup}

This section presents the set of performed experiments, the used evaluation metrics, and the method used to create the synthetic masks.

\subsection{Experiments}

Each experimental setup targets a certain comparison pair configuration, based on the existence of the mask and its type (real or synthetic) in the reference and probe images.
All experimental setups are used to evaluate the verification performance of the 4 considered automated recognition systems (Arc, Sphere, VGG, COTS), as well as the human experts under two different configuration sets (Human-c1 and Human-c2). 
As a baseline, we start by the evaluation of face verification performance without the influence of wearing a mask.
This evaluation is done by performing an N:M (all N reference images and all M probe images) comparison between the data splits BLR and BLP (BLR-BLP).
In the following step, we perform an N:M evaluation between the data subsets BLR and M12P (BLR-M12P) to measure the verification performance when the probe subject is wearing a mask.
To compare that to the effect of the real masks to that of our synthetic masks, we conduct an N:M evaluation between the data subsets BLR and SMP (BLR-SMP) 
Additionally, to evaluate the real masked-to-masked face verification performance, we performed an N:M comparison between the data splits M12R and M12P (M12R-M12P).
To analyze any different behavior if the reference was synthetically masked or both probes and references were synthetically masked, we conduct two N:M verification comparisons between SMR and M12P (SMR-M12P), and between SMR and SMP (SMR-SMP).
This leads to a total of 6 experimental configurations: BLR-BLP, BLR-M12P, BLR-SMP, M12R-M12P, SMR-M12P, SMR-SMP.
The detailed research questions answered by comparing the results of these combinations (bu human experts and automatic solutions) are presented and discussed with the support of the evaluation results in Section \ref{sec:results}.

In addition to the quantitative experiments, the human experts were asked to make comments on the relative confidence they had in making the decision in each of the experimental configurations in relation to others. Experts were asked to only make such comments when the confidence in making the decision was clear to them. These comments were considered when at least 4 of the 12 experts made the same consistent observation and no other expert made a contradictory observation. Later on, in the results (Section \ref{ssec:HEOb}), we report and discuss the main human experts' observations.

\subsection{Evaluation metrics}

We additionally report a set of verification performance metrics based on the ISO/IEC 19795-1 \cite{mansfield2006information} standard.
Face detection, as an essential pre-processing step, can be affected by the strong appearance change induced by wearing a face mask.
To capture that effect, we report the failure to extract rate (FTX) for our experiments.
FTX measures the rate of comparisons where the feature extraction was not possible and thus a template was not created.
Beyond that, and for the comparisons where the templates are generated successfully, we report a set of algorithmic verification performance metrics.
From these, we report the Equal Error Rate (EER), which is defined as the false non-match rate (FNMR) or the false match rate (FMR) at the operation point where they are equal.
From the algorithmic verification performance metrics, we additionally present the FNMR at different decision thresholds by reporting the lowest FNMR for an FMR $\leq$1.0\%, $\leq$0.1\%, and $\leq$0\%, namely the FMR100, FMR1000, and ZeroFMR , respectively.

Furthermore, we enrich our reported evaluation results by reporting the Fisher Discriminant Ratio (FDR) to provide an in-depth analysis of the separability of genuine and imposters scores for different experimental settings. This also indicates a comparison between the generalizability of a different solution if they achieved similar verification performance. FDR is a class separability criterion described by \cite{poh2004study} and it is given by:
\begin{equation*}
    FDR=\frac{(\mu_G - \mu_I)^2}{(\sigma_G)^2+(\sigma_I)^2},
\end{equation*}
where $\mu_G$ and $\mu_I$ are the genuine and imposter scores mean values and $\sigma_G$ and $\sigma_I$ are their standard deviations values.  The larger FDR value, the higher is the separation between the genuine and imposters scores and thus better verification performance.

Additionally, we plot the comparison score distributions of the different experimental setups. This enables deeper analyses of the distribution shifts caused by the different scenarios wearing a mask, on both automatic FR and human experts.
We also present the mean of imposter comparisons scores (I-mean) and the mean of the genuine comparisons scores (G-mean) for each experiment, to analyze the comparison scores shifts in a quantitative manner.
Furthermore, we report the receiver operating characteristic (ROC) curves for different experiments to illustrate the verification performance at a wider range of thresholds (operation points) for the different considered FR systems. Along with the ROC curves, we list the achieved area under the ROC curve (AUC). Higher AUC indicates a better verification performance (AUC of 1.0 indicates complete genuine/imposter separation).

\subsection{Simulated mask}
\label{sec:simulated_mask}

For our faces with the simulated masks, we use the synthetic mask generation method described by NIST report \cite{ngan2020ongoing}. The synthetic generation method depends on the Dlib toolkit \cite{DBLP:journals/jmlr/King09} to detect and extract $68$ facial landmarks from a face image. Based on the extracted landmark points, a face mask of different shapes, heights, and colors can be drawn on the face images. The detailed implementation of the synthetic generation method is described in \cite{ngan2020ongoing}. The synthetic mask generation method provided different face mask types with different high and coverage. To generate a synthetic mask database, we extract first the facial landmarks of each face image. Then, for each face image, we generate a synthetic masked image from mask type C and the color blue, described in \cite{ngan2020ongoing}. Samples of the simulated mask can be seen in Figure \ref{fig:sam:SM}.

\section{Experimental results}
\label{sec:results}

In this section, we present the verification performance achieved by the 4 automatic face verification systems and the two human expert verification setups (crops). The presentation of the results will target answering specific groups of questions based on two application scenarios. The first is related to having the typical unmasked reference (e.g. passport). The second questions the sanity of having a masked-face image as a reference. In both scenarios, we look into the following: (a) the effect of wearing a mask on the verification performance of both, software solutions and human experts and how do they compare to each other. (b) the correctness of using simulated masks to represent face masks in evaluating the verification performance of software solutions and human experts and how do they compare to each other. Additionally, we list the main observations made by the participating human experts on the verification process and discuss their correlation to the statistical performance. We conclude our evaluation with overall take-home messages of the reported results.

\begin{table*}[ht!]
\centering
\scriptsize
\setlength\tabcolsep{3pt}
\begin{tabular}{lllllllll}
\hline
\multicolumn{9}{c}{\textbf{BLR-BLP}}\\
\hline
\textbf{System} & \textbf{EER} & \textbf{FMR100} & \textbf{FMR1000} & \textbf{ZeroFMR} & \textbf{G-mean} & \textbf{I-mean} & \textbf{FDR} & \textbf{FTX}\\
\hline
Arc & 0.0000\% & 0.0000\% & 0.0000\% & 0.0000\% & 0.702020 & 0.302603 & 33.4823 & 0.000\%\\
\hline
Sphere & 2.5237\% & 3.5855\% & 9.1957\% & 43.1286\% & 0.703659 & 0.132542 & 9.2316 & 0.000\%\\
\hline
VGG & 2.3795\% & 4.6561\% & 9.9513\% & 19.2483\% & 0.819001 & 0.596985 & 8.681 & 0.000\%\\
\hline
COTS & 0.0000\% & 0.0000\% & 0.0000\% & 0.0000\% & 112.317473 & 1.630475 & 145.2178 & 0.000\%\\
\hline
Human-c1 & 0.0000\% & 0.0000\% & 0.0000\% & 0.0000\% & 9.920000 & 0.180645 & 104.6568 & --\\
\hline
Human-c2 & 0.0000\% & 0.0000\% & 0.0000\% & 0.0000\% & 9.985294 & 0.074830 & 476.3922 & --\\
\hline
\end{tabular}
\caption{The verification performance on unmasked-faces BLR-BLP of the 4 automatic face verification solutions and the human experts under two cropping scenarios. ArcFace, COTS, Human-c1, and Human-c2.}\label{eer_table1_blr_blp}
\end{table*}

\subsection{Scenario 1: verifying unmasked references to masked probes}

\noindent\textbf{As a baseline, how does the human experts' performance compare to different software solutions, in verifying unmasked-faces?}

As demonstrated in Table \ref{eer_table1_blr_blp}, ArcFace and COTS performed with no verification errors. VGGFace and SphereFace scored an EER between 2-3\%. The human experts, under both cropping setups, also performed perfectly, with the wider crop (Human-c2) producing more separable decisions than the tighter crop (Human-c1) represented by the higher FDR values. As expected from face images in a collaborative environment and with no mask, all automatic systems reported 0\% FTX, pointing out the good detectability of unmasked-faces.

Based on these results, we can conclude that the better performing automatic FR systems (COTS and ArcFace) and the human experts achieve excellent verification performance under the given database conditions. Previous works have shown that under more challenging capture scenarios (in-the-wild), automatic solutions can perform even better than human experts \cite{DBLP:conf/aaai/LuT15}. In our experiment, we notice that one of the automatic solutions (COTS) achieved a higher FDR value than the human experts, pointing out a higher genuine/imposter class separability.

\begin{table*}[ht!]
\centering
\scriptsize
\setlength\tabcolsep{3pt}
\begin{tabular}{lllllllll}
\hline
\multicolumn{9}{c}{\textbf{BLR-M12P}}\\
\hline
\textbf{System} & \textbf{EER} & \textbf{FMR100} & \textbf{FMR1000} & \textbf{ZeroFMR} & \textbf{G-mean} & \textbf{I-mean} & \textbf{FDR} & \textbf{FTX}\\
\hline
Arc & 2.8121\% & 3.3917\% & 4.2792\% & 11.0740\% & 0.490206 & 0.298841 & 9.3521 & 4.4237\%\\
\hline
Sphere & 15.6472\% & 57.7897\% & 79.1916\% & 96.8343\% & 0.280806 & 0.039197 & 1.8983 & 4.4237\%\\
\hline
VGG & 18.9208\% & 47.5859\% & 67.8120\% & 86.9508\% & 0.639760 & 0.544981 & 0.6378 & 4.4237\%\\
\hline
COTS & 1.0185\% & 1.0747\% & 1.6454\% & 5.3296\% & 67.330914 & 1.666779 & 29.9121 & 0.000\%\\
\hline
Human-c1 & 2.6405\% & 4.0000\% & 10.6666\% & 10.8114\% & 9.186667 & 0.222222 & 22.9016 & --\\
\hline
Human-c2 & 0.3448\% & 0.0000\% & 1.4705\% & 1.4829\% & 9.500000 & 0.041379 & 52.6956 & --\\
\hline
\end{tabular}
\caption{The verification performance on unmasked references and masked (real) probes BLR-M12P of the 4 automatic face verification solutions and the human experts under two cropping scenarios. In comparison to unmasked probes (Table \ref{eer_table1_blr_blp}), all automatic and human expert solutions loose verification accuracy.}\label{eer_table1_blr_m12p}
\end{table*}

\begin{figure*}[ht!]
     \centering
     \begin{subfigure}[b]{0.32\textwidth}
         \centering
         \includegraphics[width=\textwidth]{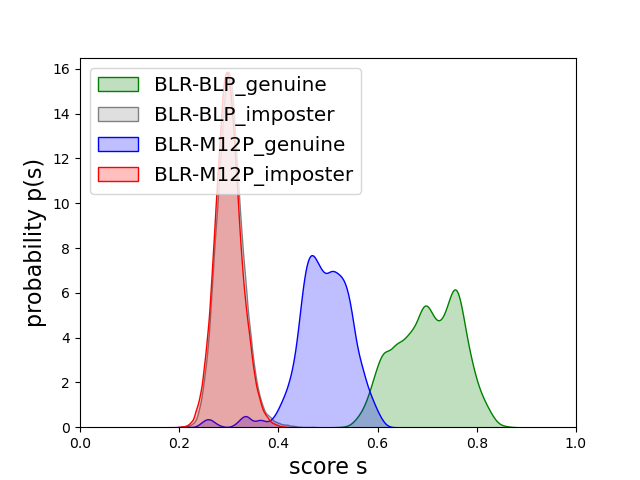}
         \caption{Arc}
         \label{fig:samp:arc1}
     \end{subfigure}
     \hfill
     \begin{subfigure}[b]{0.32\textwidth}
         \centering
         \includegraphics[width=\textwidth]{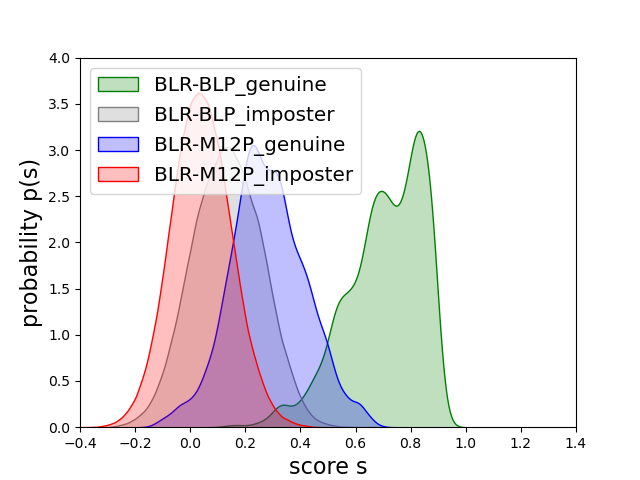}
         \caption{Sphere}
         \label{fig:sam:sphere2}
     \end{subfigure}
     \hfill
     \begin{subfigure}[b]{0.32\textwidth}
         \centering
         \includegraphics[width=\textwidth]{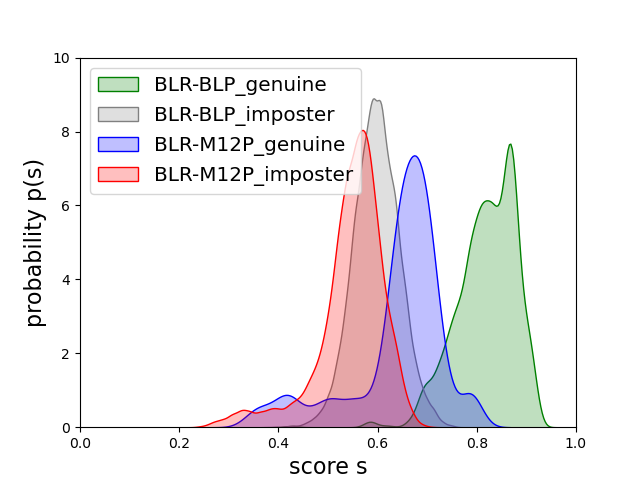}
         \caption{VGG}
         \label{fig:sam:vgg3}
     \end{subfigure}
     \hfill
     \begin{subfigure}[b]{0.32\textwidth}
         \centering
         \includegraphics[width=\textwidth]{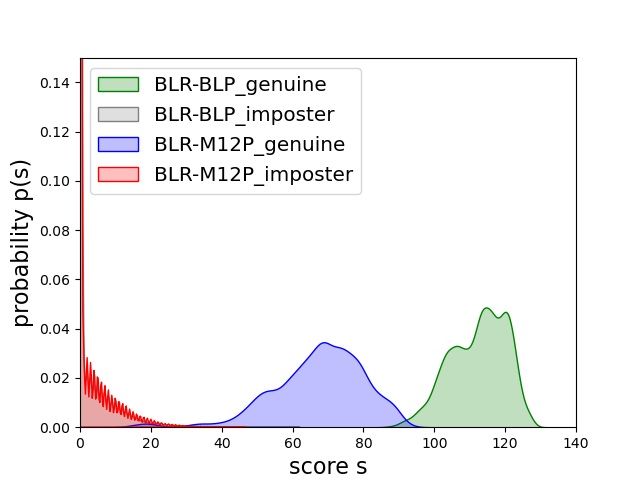}
         \caption{COTS}
         \label{fig:sam:cots4}
    \end{subfigure}    
    \begin{subfigure}[b]{0.32\textwidth}
         \centering
         \includegraphics[width=\textwidth]{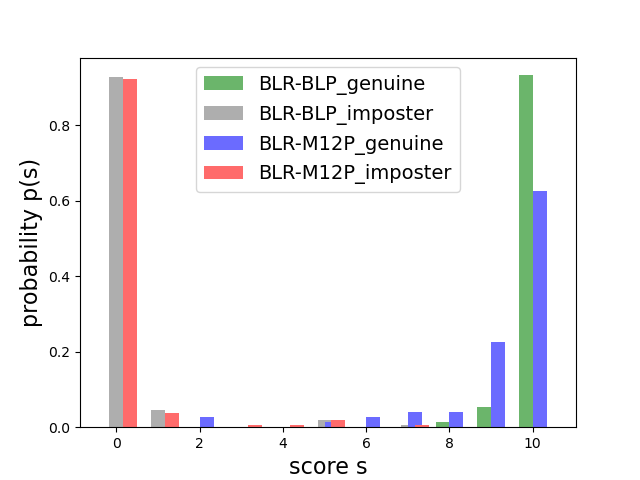}
         \caption{Human-c1}
         \label{fig:sam:hc15}
     \end{subfigure}
     \hfill
     \begin{subfigure}[b]{0.32\textwidth}
         \centering
         \includegraphics[width=\textwidth]{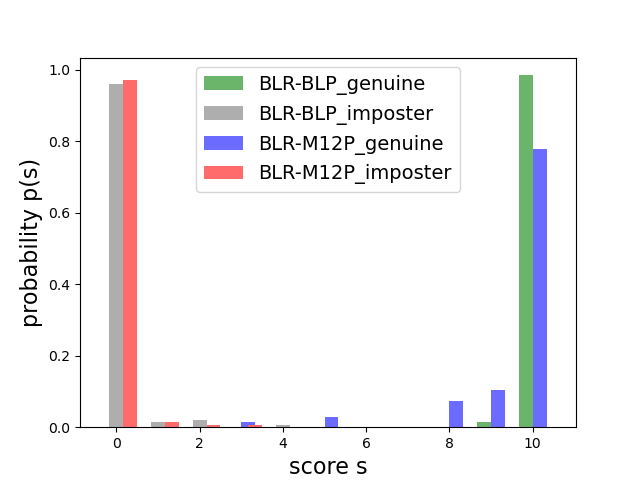}
         \caption{Human-c2}
         \label{fig:sam:hc26}
     \end{subfigure}
     \caption{The comparison score (similarity) distributions comparing the "baseline" BLR-BLP genuine and imposter distributions to those of the distributions including "masked" faces probes (BLR-M12P). The shift of the genuine scores towards the imposter distribution is clear when faces are masked for all investigated automatic solutions,and to a lower degree for human experts. Imposter scores distributions is relatively less effected.}
    \label{fig:dist_blrblp_blrm12p}
     \hfill
\end{figure*}

\begin{figure*}[ht!]
     \centering
      \includegraphics[width=0.63\textwidth]{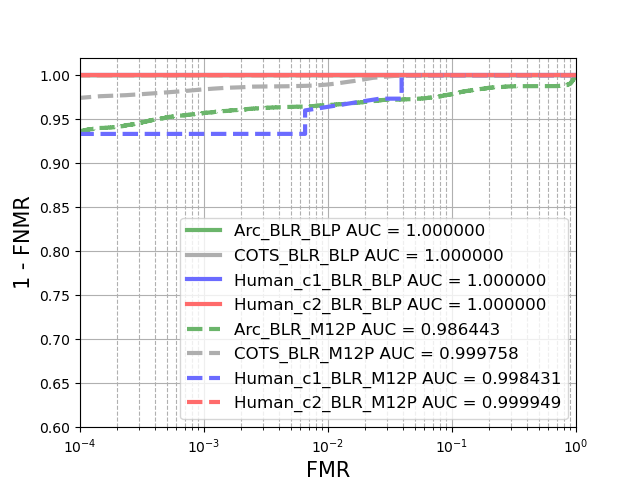}
      \caption{ROC curves of the ArcFace, COTS, Human-c1, Human-c2 on the BLR-BLP and BLR-M12P configurations. All setups loose performance when dealing with Masked probes (M12P).}
      \label{fig:roc_BLP_vs_M12P}
\end{figure*}

\noindent\textbf{In comparison to the unmasked baseline, how does human experts performance compare to software solutions, in verifying masked probes to unmasked references?}

The verification performances achieved when an unmasked reference is compared to a real masked probe (BLR-M12P) are presented in Table \ref{eer_table1_blr_m12p}. In comparison to the baseline (BLR-BLP) in Table \ref{eer_table1_blr_blp}, the performances of all automatic solutions and human experts are affected by the masked-face probe. SphereFace and VGGFace moved from an EER of under 3\% to an EER over 15\%. The EER of the better performing ArcFace and COTS moved up from 0\% in the BLR-BLP to 2.8\% and 1.0\%, respectively, when dealing with masked probes. The performance of human experts was also affected by the masked probe with the EER values moving from 0\% in the BLR-BLP scenario to 2.6\% and 0.34\%, respectively for Human-c1 and Human-c2. Here, the wider crop achieved significantly better verification performance, indicating that some of the experts' decisions were affected by properties on the border of the face, such as the hairstyle, which might not be realistic in long-term verification.
Similar conclusions can be made in the demonstrated ROC curves for the BLR-BLP and BLR-M12P setups of the top-performing automatic solutions (ArcFace and COTS) and the two human expert setups, see Figure \ref{fig:roc_BLP_vs_M12P}. 
The FTX values in Table \ref{eer_table1_blr_m12p} indicate that the investigated academic automatic systems, all using the same face detector (MTCNN), have failed to detect the masked-faces in some cases, leading to an FTX value of 4.4\%.

Looking at the distributions of the genuine and imposter scores of all verification solutions in Figure \ref{fig:dist_blrblp_blrm12p}, one can notice a shift in the genuine scores towards the imposter scores when comparing the BLR-BLP to the BLR-M12P setups, i.e. when the probes are masked. The shift in the imposter scores towards the genuine ones was smaller and almost negligible in the better performing ArcFace, COTS, and the two human expert setups. These remarks can also be concluded by comparing the I-mean and G-mean values in Tables \ref{eer_table1_blr_blp} and \ref{eer_table1_blr_m12p}.

Based on the presented results, we can point out the consistent negative effect of having a masked probe on the verification performance of the best performing automatic solutions and the human experts. It is noted as well that most of this effect, even for human experts, was induced by a shift in the genuine score values, which corresponds to higher FNMR. Looking at the human experts' verification performance and FDR under both cropping scenarios, we can notice that Human-c2 performed better than COTS (best automatic solutions), while Human-c1 performed worse than COTS.

\begin{table*}[ht!]
\centering
\scriptsize
\setlength\tabcolsep{4pt}
\begin{tabular}{lllllllll}
\hline
\multicolumn{9}{c}{\textbf{BLR-SMP}}\\
\hline
\textbf{System} & \textbf{EER} & \textbf{FMR100} & \textbf{FMR1000} & \textbf{ZeroFMR} & \textbf{G-mean} & \textbf{I-mean} & \textbf{FDR} & \textbf{FTX}\\
\hline
Arc & 1.1651\% & 1.3001\% & 5.4550\% & 13.0723\% & 0.507047 & 0.300292 & 12.3798 & 5.114\%\\
\hline
Sphere & 15.4323\% & 31.6845\% & 46.3397\% & 76.9502\% & 0.415802 & 0.48133 & 2.6162 & 5.114\%\\
\hline
VGG & 7.6525\% & 14.1156\% & 54.2517\% & 70.4081\% & 0.695138 & 0.576018 & 3.0399. & 5.114\%\\
\hline
COTS & 0.6002\% & 0.6337\% & 0.6337\% & 0.8684\% & 74.317568 & 1.756565 & 30.6757 & 0.000\%\\
\hline
Human-c1 & 1.6342\% & 5.4794\% & 5.4794\% & 5.4794\% & 9.315068 & 0.183544 & 24.9824 & --\\
\hline
Human-c2 & 0.0000\% & 0.0000\% & 0.0000\% & 0.0000\% & 9.732394 & 0.020548 & 114.833 & --\\
\hline
\end{tabular}
\caption{The verification performance on unmasked references and masked (synthetic) probes BLR-SMP of the 4 automatic face verification solutions and the human experts under two cropping scenarios. In comparison to unmasked probes (Table \ref{eer_table1_blr_blp}) and real masked probes (Table \ref{eer_table1_blr_m12p}), most automatic and human expert solutions loose verification accuracy however to a significantly lower degree than when the probe mask is real.}\label{eer_BLR_SMP}
\end{table*}

\begin{figure*}[ht!]
     \centering
      \includegraphics[width=0.63\textwidth]{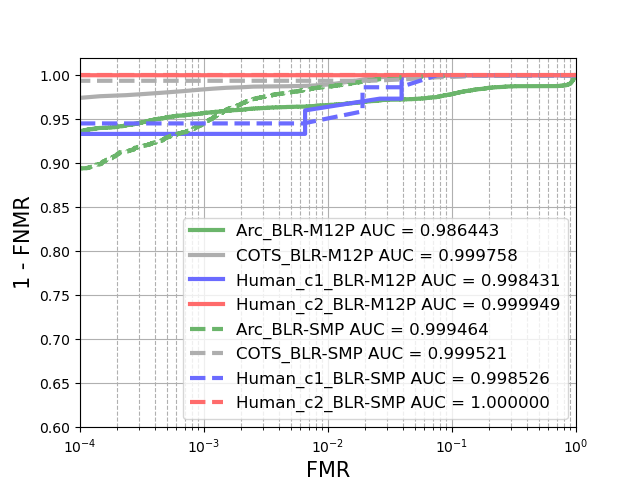}
      \caption{ROC curves of the ArcFace, COTS, Human-c1, Human-c2 on the BLR-M12P and BLR-SMP configurations. Most setups performance better when dealing with probes with simulated masks (SMP), in comparison to real masks.}
      \label{fig:roc:realmaskedprobe_Vs_simulatedmaskedprobe}
\end{figure*}

\begin{figure*}[ht!]
     \centering
     \begin{subfigure}[b]{0.32\textwidth}
         \centering
         \includegraphics[width=\textwidth]{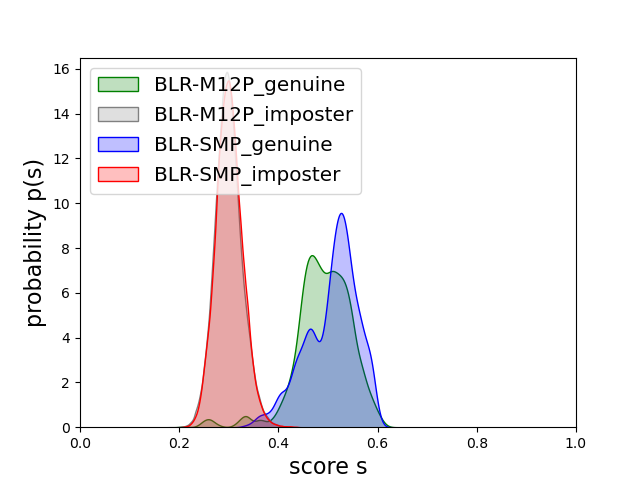}
         \caption{Arc}
         \label{fig:samp:arc11}
     \end{subfigure}
     \hfill
     \begin{subfigure}[b]{0.32\textwidth}
         \centering
         \includegraphics[width=\textwidth]{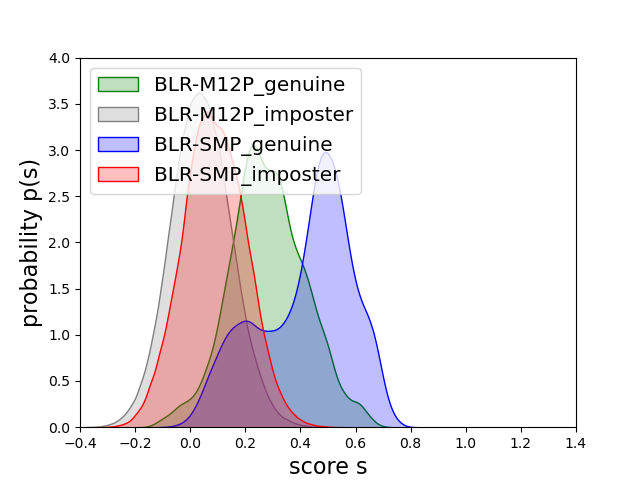}
         \caption{Sphere}
         \label{fig:sam:sphere11}
     \end{subfigure}
     \hfill
     \begin{subfigure}[b]{0.32\textwidth}
         \centering
         \includegraphics[width=\textwidth]{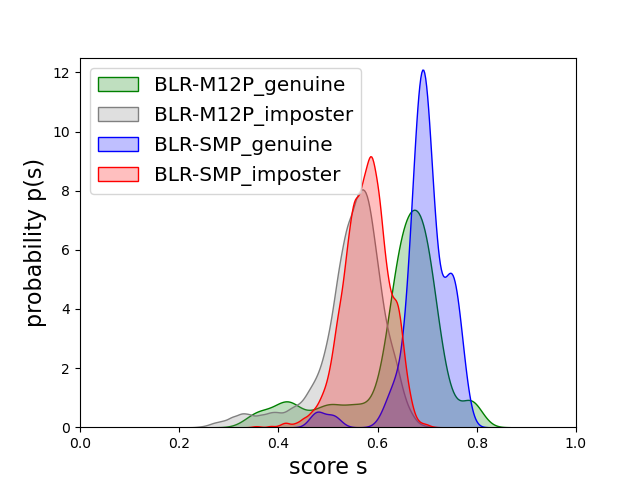}
         \caption{VGG}
         \label{fig:sam:vgg11}
     \end{subfigure}
     \hfill
     \begin{subfigure}[b]{0.32\textwidth}
         \centering
         \includegraphics[width=\textwidth]{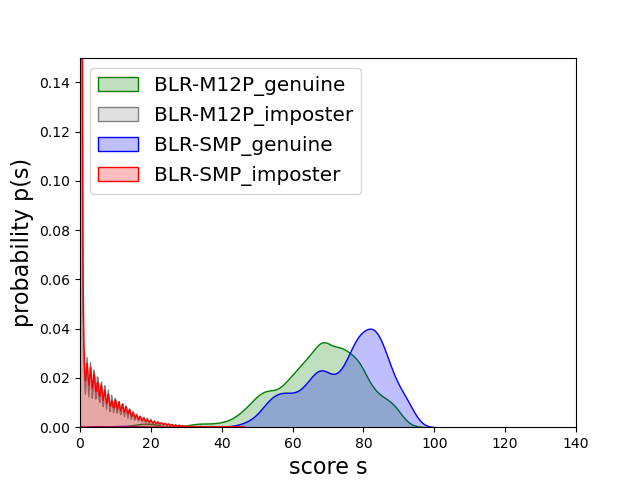}
         \caption{COTS}
         \label{fig:sam:cots11}
    \end{subfigure}    
    \begin{subfigure}[b]{0.32\textwidth}
         \centering
         \includegraphics[width=\textwidth]{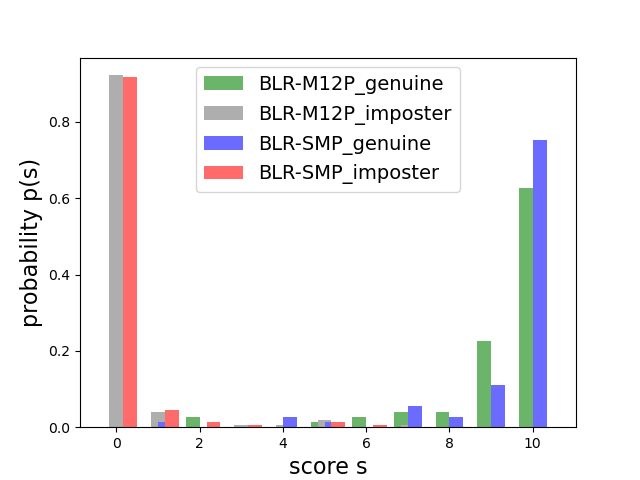}
         \caption{Human-c1}
         \label{fig:sam:hc1}
     \end{subfigure}
     \hfill
     \begin{subfigure}[b]{0.32\textwidth}
         \centering
         \includegraphics[width=\textwidth]{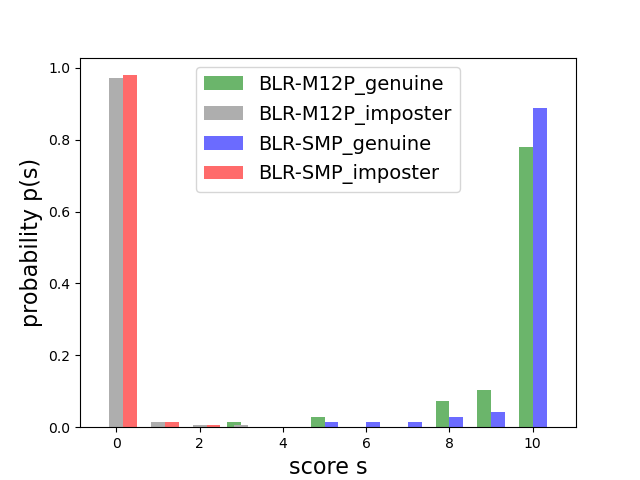}
         \caption{Human-c2}
         \label{fig:sam:hc211}
     \end{subfigure}
     \caption{The comparison score (similarity) distributions comparing the BLR-M12P genuine and imposter distributions to those of the distributions including "synthetically masked" faces probes (BLR-SMP). The shift of the genuine scores towards the imposter distribution is larger when dealing with real masked-face probes (M12P) for all investigated automatic solutions and the human expert configurations. }
    \label{fig:dist_blrm12p_blrsmp}
     \hfill
\end{figure*}

\noindent\textbf{How well can a probe with a simulated mask represent the real mask scenario in the evaluation of human expert and software solution verification performance?}

The verification performances achieved when an unmasked reference is compared to a synthetically masked probe (BLR-SMP) are presented in Table \ref{eer_BLR_SMP}. 
In comparison to the performances achieved when the probes were wearing a real mask (BLR-M12P) in Table \ref{eer_table1_blr_m12p}, the performances of all automatic solutions and human experts are less affected by the synthetically masked-face probe. 
SphereFace and VGGFace moved from an EER of 15.6\% and 18.9\% to 15.4\% and 7.7\%, respectively when probes had real masks (M12P) and when they had simulated masks(SMP). The EER of the better performing ArcFace and COTS moved down from 2.8\% and 1.0\% in the BLR-M12P to 1.2\% and 0.6\%, respectively, when dealing with simulated mask probes (BLR-SMP). The performance of human experts was also less affected by the synthetically masked probe (in comparison to real masked probe) with the EER values moving from 2.6\% and 0.3\% in the BLR-M12P scenario to 1.6\% and 0.0\%, respectively for Human-c1 and Human-c2. 
Also, here in the BLR-SMP scenario, as in the BLR-M12P one, the wider crop achieved significantly better verification performance. This again points out the significance of areas bordering the face in the human verification decision.
The FTX values when dealing with synthetic masks (Table \ref{eer_BLR_SMP}) is only slightly higher than that of real masks (Table \ref{eer_table1_blr_m12p}).
Looking at the ROC curves produced for the BLR-M12P and BLR-SMP configurations by the top-performing automatic solutions (ArcFace and COTS) and the two human expert setups (see Figure \ref{fig:roc:realmaskedprobe_Vs_simulatedmaskedprobe}), one can notice that, for all solution and human expert setups, the verification performance on most operation points was higher for synthetically masked probes (SMP), in comparison to real masked probes (M12P). This indicates that simulated masks might not be optimal to evaluate the performance of automatic FR systems, as well as the verification ability of human experts.
It would be a challenging task to simulate the realistic variations in real masks and their effects (shadow, reflection, etc.) to perform such evaluation.

The genuine and imposter scores distributions of all verification solutions in Figure \ref{fig:dist_blrm12p_blrsmp}, one can notice a shift in the genuine scores towards the imposter scores when comparing the BLR-M12P to the BLR-SMP setups, i.e. when the masks are real (vs. synthetic). 
The shift in the imposter scores towards the genuine ones was again smaller and almost negligible in the better performing ArcFace, COTS, and the two human expert setups. similar observations can be made by comparing the I-mean and G-mean values in Tables \ref{eer_table1_blr_m12p} and \ref{eer_BLR_SMP}.

Based on the presented results, we can point out that there is a clear difference between the effect of having real masked probes and having synthetically masked probes. At least for our implemented synthetic masks, the performance degradation was lower. This was the consistent observation for the automatic FR systems and the human experts' verification results.

\subsection{Scenario 2: verifying masked references to masked probes}

\noindent\textbf{In comparison to the unmasked baseline, how does the performance of human experts compare to software solutions, in verifying masked probes to masked references?}

\begin{figure*}[ht!]
     \centering
      \includegraphics[width=0.63\textwidth]{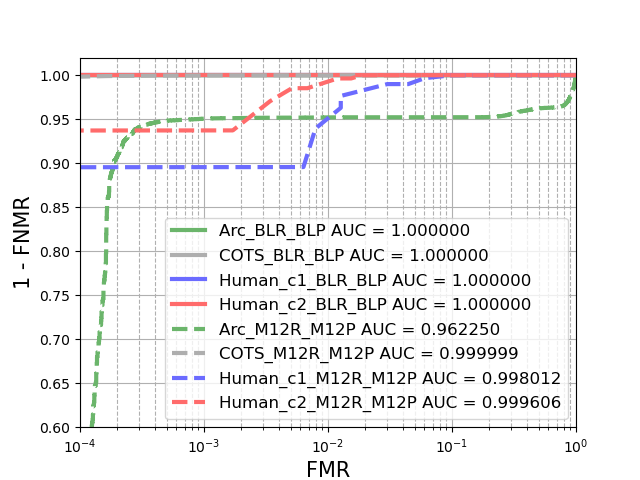}
      \caption{ROC curves of the ArcFace, COTS, Human-c1, Human-c2 on the BLR-BLP and M12R-M12P configurations. All setups loose performance better when when the comparison is between masked-faces.}
      \label{fig:maskedmasked_vs_unmaskedunmasked}
\end{figure*}

The verification performances achieved when a real masked reference is compared to a real masked probe (M12R-M12P) are presented in Table \ref{eer_M12R-M12P}. In comparison to the baseline (BLR-BLP) in Table \ref{eer_table1_blr_blp}, the performances of all automatic solutions and human experts degrade with the masks. 
SphereFace and VGGFace moved from an EER of under 3\% to an EER over 20\%. The EER of the better performing ArcFace and COTS moved up from 0\% in the BLR-BLP to 4.7\% and 0.04\%, respectively, when dealing with masked probes. The performance of human experts was also affected by the masked probe with the EER values moving from 0\% in the BLR-BLP scenario to 1.8\% and 0.78\%, respectively for Human-c1 and Human-c2. Again showing the effect of a wider crop on human expert verification decisions. 
One can notice that the EER of the COTS was only slightly affected, however, the FDR value was degraded from 145.2 in the BLR-BLP case to 37.8 in the M12R-M12P case, pointing out a large difference in the class separability.
Similar conclusions can be made, on a wider operation range, in the demonstrated ROC curves for the BLR-BLP and M12R-M12P setups of the top-performing automatic solutions (ArcFace and COTS) and the two human expert setups, see Figure \ref{fig:maskedmasked_vs_unmaskedunmasked}. 
The FTX values in Table \ref{eer_M12R-M12P} indicate that the investigated academic automatic systems, using the MTCNN detection, have failed to detect the masked-faces in some references and probes, leading to an FTX value of 4.7\%.

\begin{figure*}[ht!]
     \centering
     \begin{subfigure}[b]{0.32\textwidth}
         \centering
         \includegraphics[width=\textwidth]{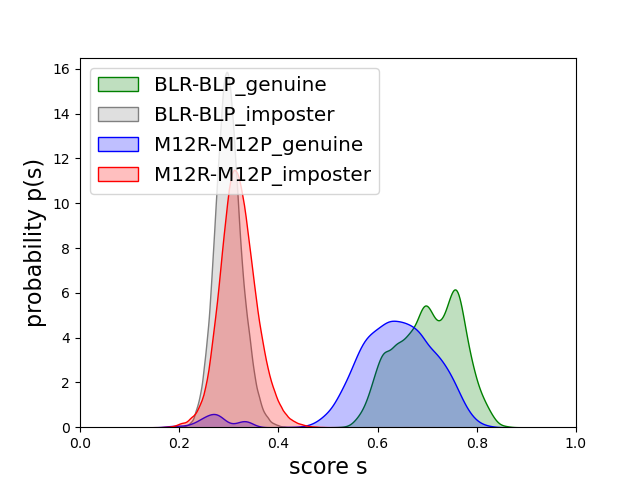}
         \caption{Arc}
         \label{fig:samp:arc111}
     \end{subfigure}
     \hfill
     \begin{subfigure}[b]{0.32\textwidth}
         \centering
         \includegraphics[width=\textwidth]{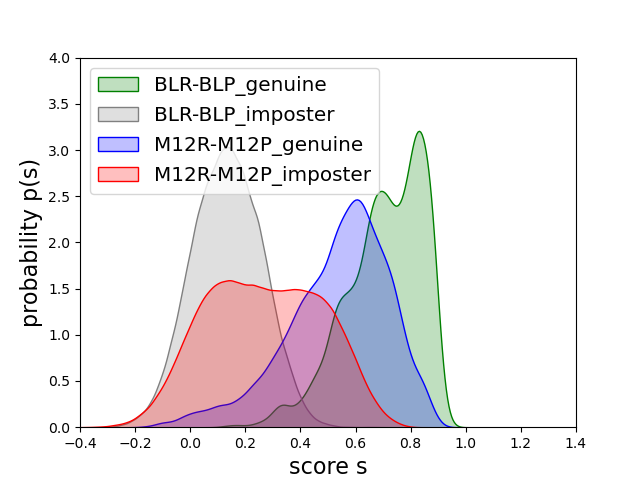}
         \caption{Sphere}
         \label{fig:sam:sphere111}
     \end{subfigure}
     \hfill
     \begin{subfigure}[b]{0.32\textwidth}
         \centering
         \includegraphics[width=\textwidth]{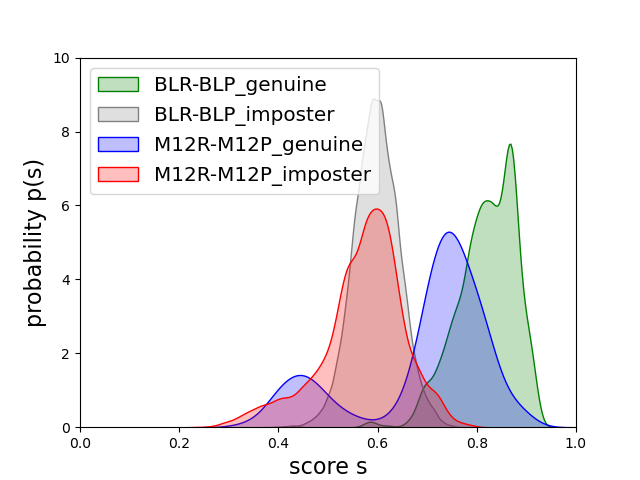}
         \caption{VGG}
         \label{fig:sam:vgg111}
     \end{subfigure}
     \hfill
     \begin{subfigure}[b]{0.32\textwidth}
         \centering
         \includegraphics[width=\textwidth]{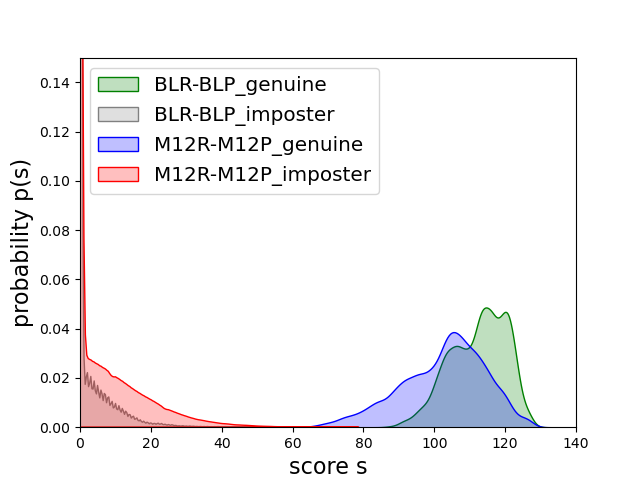}
         \caption{COTS}
         \label{fig:sam:cots111}
    \end{subfigure}    
    \begin{subfigure}[b]{0.32\textwidth}
         \centering
         \includegraphics[width=\textwidth]{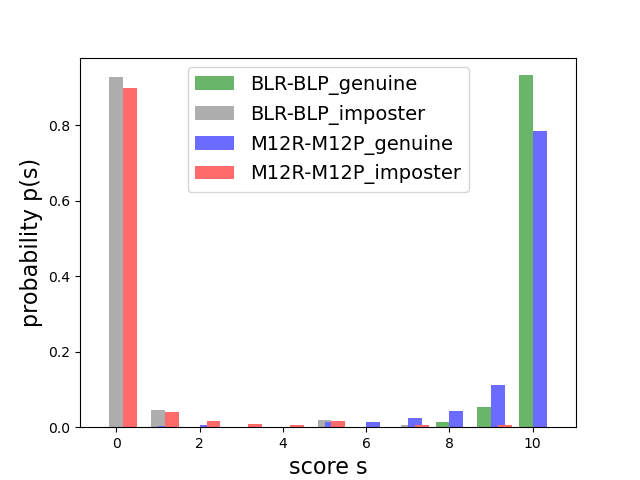}
         \caption{Human-c1}
         \label{fig:sam:hc1111}
     \end{subfigure}
     \hfill
     \begin{subfigure}[b]{0.32\textwidth}
         \centering
         \includegraphics[width=\textwidth]{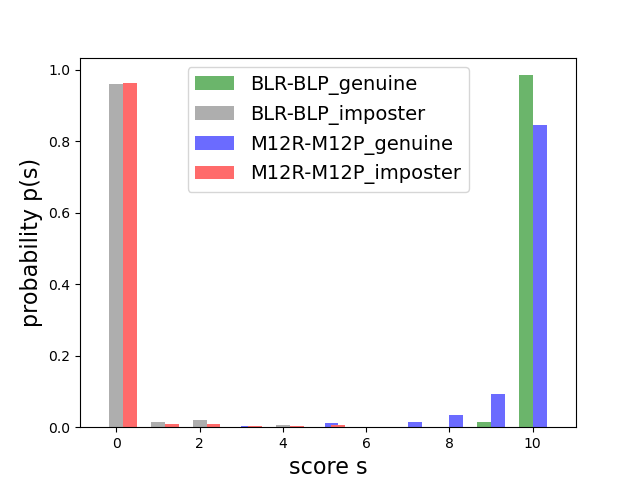}
         \caption{Human-c2}
         \label{fig:sam:hc2111}
     \end{subfigure}
     \caption{The comparison score (similarity) distributions comparing the "baseline" BLR-BLP genuine and imposter distributions to those of the distributions including masked (real) faces probes and references (M12R-M12P). The shift of the genuine scores towards the imposter distribution is clear when faces are masked for all investigated automatic solutions,and human experts. Imposter scores distributions is relatively less effected, however, relatively (to genuine scores shift) more effected than the case where only the probes are masked (Figure \ref{fig:dist_blrblp_blrm12p}).}
    \label{fig:dist_blrblp_m12rm12p}
\end{figure*}

The genuine and imposter scores distributions of the M12R-M12P in comparison to BLR-BLP are shown in Figure \ref{fig:dist_blrblp_m12rm12p}. The figure shows a shift in genuine and imposter scores towards each other when the references and probes are masked. The shift of the genuine scores towards the imposter range seems to be clearer than that of the imposter scores towards the genuine range, especially in the better performing ArcFace, COTS, and both human experts settings. This also can be noted when comparing the I-mean and G-mean values in Tables \ref{eer_table1_blr_blp} and \ref{eer_M12R-M12P}. One must note that the relative shift in the imposter scores was smaller in the case of BLR-M12P, where the genuine shift was larger (see Figure \ref{fig:dist_blrblp_blrm12p}).

Based on the presented results, both human experts and automatic FR solutions lose performance when verifying two masked-faces, in comparison to verifying unmasked-faces. The human performance in this earlier case is lower than the best performing automatic solution and better than the worse 4 automatic solutions. Also in this experiment, the wider crop proved to enhance the human expert performance on such a database that does not include large (years) time gaps between the reference and probe sessions.

\begin{table*}[ht!]
\centering
\scriptsize
\setlength\tabcolsep{3pt}
\begin{tabular}{lllllllll}
\hline
\multicolumn{9}{c}{\textbf{M12R-M12P}}\\
\hline
\textbf{System} & \textbf{EER} & \textbf{FMR100} & \textbf{FMR1000} & \textbf{ZeroFMR} & \textbf{G-mean} & \textbf{I-mean} & \textbf{FDR} & \textbf{FTX}\\
\hline
Arc & 4.7809\% & 4.7959\% & 4.9427\% & 99.7933\% & 0.624205 & 0.318468 & 7.6235 & 4.736\%\\
\hline
Sphere & 25.5317\% & 78.5115\% & 91.3690\% & 99.5655\% & 0.533365 & 0.266994 & 0.9161\ & 4.736\%\\
\hline
VGG & 20.9248\% & 51.3110\% & 76.9665\% & 84.9871\% & 0.693083 & 0.569062 & 0.6135 & 4.736\%\\
\hline
COTS & 0.0417\% & 0.0000\% & 0.0248\% & 3.2992\% & 102.408421 & 6.735984 & 37.8482 & 0.000\%\\
\hline
Human-c1 & 1.8123\% & 6.0606\% & 21.5488\% & 21.5488\% & 9.525253 & 0.310618 & 28.381 & --\\
\hline
Human-c2 & 0.7817\% & 3.6900\% & 6.2730\% & 15.4981\% & 9.715867 & 0.126280 & 67.2575 & --\\
\hline
\end{tabular}
\caption{The verification performance on masked (real) references and masked (real) probes  M12R-M12P of the 4 automatic face verification solutions and the human experts under two cropping scenarios. In comparison to unmasked baseline (Table \ref{eer_table1_blr_blp}), all automatic and human expert solutions loose verification accuracy. In comparison to having an unmasked references and a masked (real) probe (Table \ref{eer_table1_blr_m12p}), the lower performing academic FR solution loose performance, while the top performing COTS and human experts perform better.}\label{eer_M12R-M12P}
\end{table*}

\noindent\textbf{Is it worth it to have a masked reference? how does verifying masked probes to masked references compare to verifying them to unmasked references, for human experts and software solutions?}

It might be intuitive to opt to compare two masked-faces rather than a masked-face to an unmasked-face. This can be driven by the idea that an algorithm or human expert will put more focus on comparing the shared, uncovered area. Therefore, it might be useful to have a masked, whether real or simulated, reference of the user. The following experimental results will address this open question.

\begin{figure*}[ht!]
     \centering
      \includegraphics[width=0.63\textwidth]{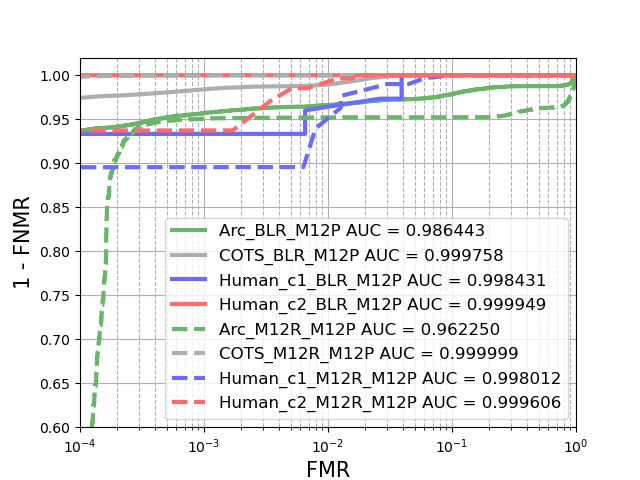}
      \caption{ROC curves of the ArcFace, COTS, Human-c1, Human-c2 on the BLR-M12P and M12R-M12P configurations. Different setups at different operation points have mixed relative performances when comparing two masked-faces in comparison to compering unmasked-to-masked faces. }
      \label{fig:roc_maskedmasked_vs_maskedunmasked}
\end{figure*}

The verification performances achieved when a real masked reference is compared to a real masked probe (M12R-M12P) are presented in Table \ref{eer_M12R-M12P}. In comparison to comparing unmasked references to masked probes (BLR-M12P) in Table \ref{eer_table1_blr_m12p}, the performances enhance in the M12R-M12P setting for the top-performing automatic solutions COTS and the Human-c1 setting. The separability, given as FDR, also increases for COTS and all the human expert settings. However, for the less performing ArcFace, SphereFace, and VGGFace, the verification performances and separability decrease in the case of M12R-M12P in comparison to BLR-M12P.
Similar conclusions can be made in the demonstrated ROC curves for the BLR-M12P and M12R-M12P setups of the top-performing automatic solutions (ArcFace and COTS) and the two human expert setups, see Figure \ref{fig:roc_maskedmasked_vs_maskedunmasked}. The figure shows that different setups at different operation points have mixed relative performances when comparing two masked-faces in comparison to comparing unmasked-to-masked faces.

\begin{figure*}[ht!]
     \centering
     \begin{subfigure}[b]{0.32\textwidth}
         \centering
         \includegraphics[width=\textwidth]{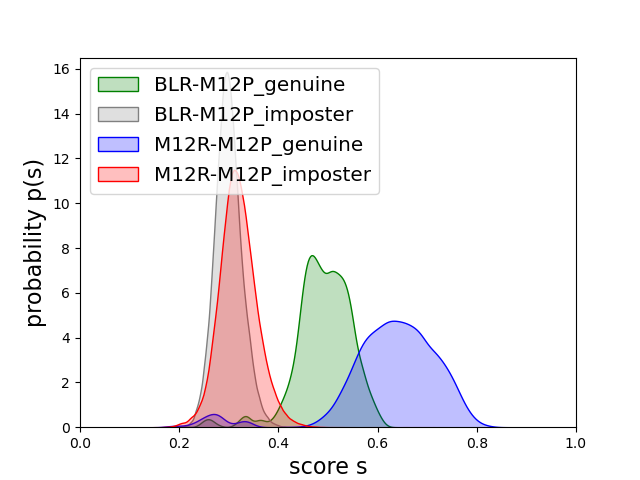}
         \caption{Arc}
         \label{fig:samp:arc22}
     \end{subfigure}
     \hfill
     \begin{subfigure}[b]{0.32\textwidth}
         \centering
         \includegraphics[width=\textwidth]{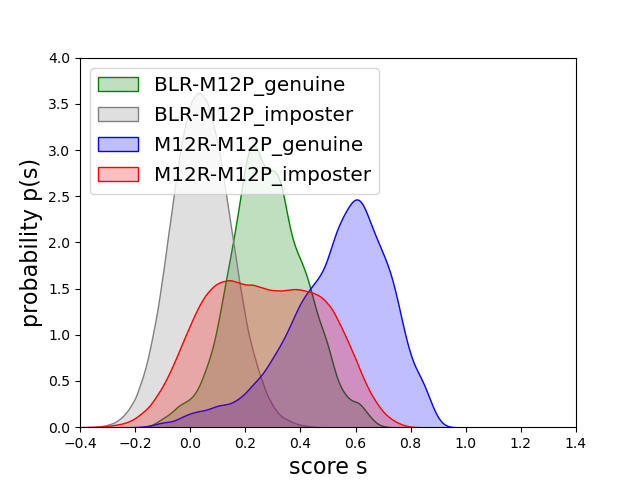}
         \caption{Sphere}
         \label{fig:sam:sphere22}
     \end{subfigure}
     \hfill
     \begin{subfigure}[b]{0.32\textwidth}
         \centering
         \includegraphics[width=\textwidth]{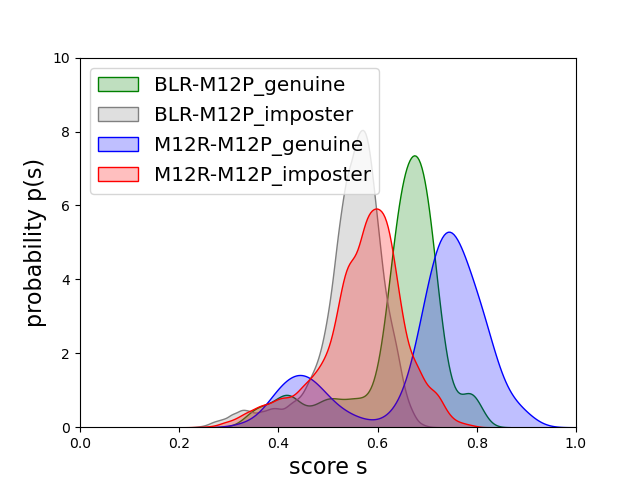}
         \caption{VGG}
         \label{fig:sam:vgg22}
     \end{subfigure}
     \hfill
     \begin{subfigure}[b]{0.32\textwidth}
         \centering
         \includegraphics[width=\textwidth]{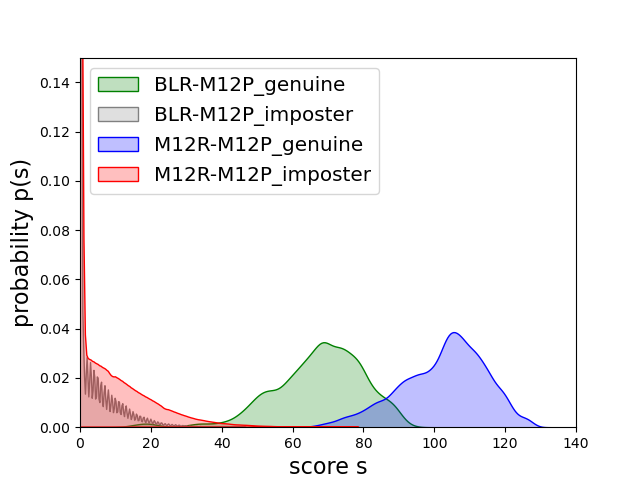}
         \caption{COTS}
         \label{fig:sam:cots22}
    \end{subfigure}    
    \begin{subfigure}[b]{0.32\textwidth}
         \centering
         \includegraphics[width=\textwidth]{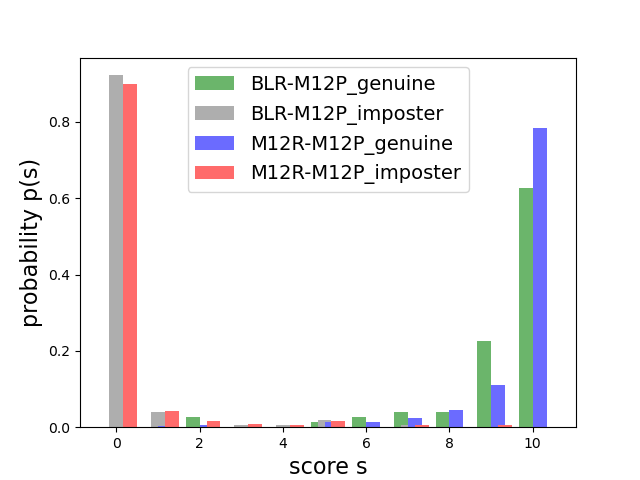}
         \caption{Human-c1}
         \label{fig:sam:hc122}
     \end{subfigure}
     \hfill
     \begin{subfigure}[b]{0.32\textwidth}
         \centering
         \includegraphics[width=\textwidth]{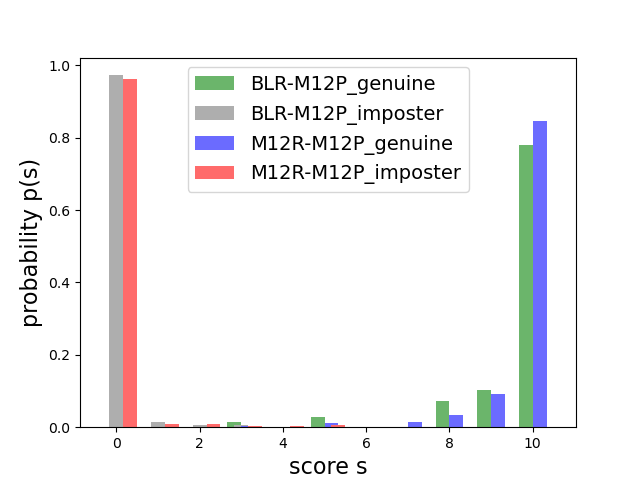}
         \caption{Human-c2}
         \label{fig:sam:hc222}
     \end{subfigure}
     \caption{The comparison score (similarity) distributions comparing the BLR-M12P genuine and imposter distributions to those of the distributions including masked (real) faces probes and references (M12R-M12P). The shift of the genuine scores towards the imposter distribution is slightly larger, in most cases, when both compared faces are masked (M12R-M12P).}
    \label{fig:dts_blrm12p_m12rm12p}
     \hfill
\end{figure*}

The genuine and imposter scores distributions of the M12R-M12P in comparison to BLR-M12P are shown in Figure \ref{fig:dts_blrm12p_m12rm12p}. The figure shows a larger shift in the BLR-M12P genuine scores towards the imposter range, in comparison to the M12R-M12P. It also shows a larger shift in the M12R-M12P imposter scores towards the genuine range in comparison to the BLR-M12P. This is the trend in all automatic FR systems and the two human expert settings. This is also demonstrated by the comparing the I-mean and G-mean values in the Tables \ref{eer_M12R-M12P} and \ref{eer_table1_blr_m12p}.

So far, having a masked-face reference to compare a masked probe to it has produced mixed results. It is also noted that it is practically difficult to include such a second reference capture in the current biometric systems (e.g. second passport picture). Therefore, we investigate a more practical solution, which is to synthetically add a mask to the existing reference image. To investigate that, we evaluate the verification performances on the SMR-M12P configuration and compare it to the BLR-M12P one. The achieved performances of these configurations are seen in Tables \ref{eer_SMR_M12P} and \ref{eer_table1_blr_m12p} respectively. This comparison results in a clearer trend, all automatic systems (other than the least performing VGGFace) produce lower FMR100 in the SMR-M12P configuration, in comparison to the BLR-M12P. The same comparison shows an increase in the FDR separability measure on all automatic and human experts setups in the SMR-M12P configuration. Both the G-mean and I-mean values have increased in the SMR-M12P in comparison to the BLR-M12P configuration.

Based on the presented results, having a real masked reference, rather than an unmasked-face, might not enhance the verification performance of either automatic solutions or human experts. This might be due to the fact that, on one hand, similar masks in imposter pairs might lead to higher similarity. On the other hand, different masks in genuine pairs might lead to lower similarity. However, this effect is lower when using simulated masks because the reference synthetic mask is always of the same style. This is demonstrated by the consistent signs in our reported results that having a synthetically masked reference might enhance the performance in comparison to comparing masked probes with unmasked references.

\begin{table*}[ht!]
\centering
\scriptsize
\setlength\tabcolsep{3pt}
\begin{tabular}{lllllllll}
\hline
\multicolumn{9}{c}{\textbf{SMR-M12P}}\\
\hline
\textbf{System} & \textbf{EER} & \textbf{FMR100} & \textbf{FMR1000} & \textbf{ZeroFMR} & \textbf{G-mean} & \textbf{I-mean} & \textbf{FDR} & \textbf{FTX}\\
\hline
Arc & 3.3773\% & 3.3774\% & 3.6432\% & 27.4719\% & 0.606401 & 0.320682 & 9.7279.0 & 5.178\%\\
\hline
Sphere & 12.0662\% & 42.4475\% & 72.3555\% & 93.4392\% & 0.402116 & 0.053940 & 2.752 & 5.178\%\\
\hline
VGG & 23.3173\% & 53.4562\% & 70.2304\% & 82.1198\% & 0.684729 & 0.572444 & 0.7245 & 5.178\%\\
\hline
COTS & 0.2873\% & 0.0000\% & 0.6416\% & 2.7484\% & 97.883362 & 5.859622 & 39.084 & 0.000\%\\
\hline
Human-c1 & 1.6684\% & 5.5555\% & 18.0555\% & 18.0555\% & 9.638889 & 0.344156 & 30.3476 & --\\
\hline
Human-c2 & 0.0000\% & 0.0000\% & 0.0000\% & 0.0000\% & 9.716418 & 0.085714 & 112.7123 & --\\
\hline
\end{tabular}
\caption{The verification performance on synthetically masked references and masked (real) probes  SMR-M12P of the 4 automatic face verification solutions and the human experts under two cropping scenarios. In comparison to real masked references and probes (M12R-M12P) (Table \ref{eer_M12R-M12P}), all automatic and human expert solutions perform better. In comparison to having an unmasked references and a masked (real) probe BLR-M12P (Table \ref{eer_table1_blr_m12p}), most of the solutions perform better or in the same range, except ArcFace.}\label{eer_SMR_M12P}
\end{table*}

\begin{table*}[ht!]
\centering
\scriptsize
\setlength\tabcolsep{3pt}
\begin{tabular}{lllllllll}
\hline
\multicolumn{9}{c}{\textbf{SMR-SMP}}\\
\hline
\textbf{System} & \textbf{EER} & \textbf{FMR100} & \textbf{FMR1000} & \textbf{ZeroFMR} & \textbf{G-mean} & \textbf{I-mean} & \textbf{FDR} & \textbf{FTX}\\
\hline
Arc & 0.2732\% & 0.0000\% & 0.6150\% & 6.9195\% & 0.675107 & 0.330339 & 6.4712 & 5.371\%\\
\hline
Sphere & 12.7381\% & 23.9176\% & 39.1283\% & 68.7882\% & 0.55718 & 0.343760 & 2.332 & 5.371\%\\
\hline
VGG & 3.6799\% & 5.8282\% & 23.0061\% & 42.9447\% & 0.800212 & 0.610588 & 1.3912 & 5.371\%\\
\hline
COTS & 0.9322\% & 0.1081\% & 0.8917\% & 1.4317\% & 104.7186 & 5.8493 & 45.3178 & 0.000\%\\
\hline
Human-c1 & 2.2502\% & 10.5263\% & 23.6842\% & 23.6842\% & 9.539474 & 0.484076 & 22.6224 & --\\
\hline
Human-c2 & 1.4049\% & 4.3478\% & 4.3478\% & 4.3478\% & 9.608696 & 0.108844 & 33.2046 & --\\
\hline
\end{tabular}
\caption{The verification performance on synthetically masked references and synthetically masked (real) probes  SMR-SMP of the 4 automatic face verification solutions and the human experts under two cropping scenarios. In comparison to real masked references and probes (M12R-M12P) (Table \ref{eer_M12R-M12P}), all automatic solutions perform better with the exception of COTS (similarly to human experts). }\label{eer_SMR_SMP}
\end{table*}

\begin{figure*}[ht!]
     \centering
      \includegraphics[width=0.63\textwidth]{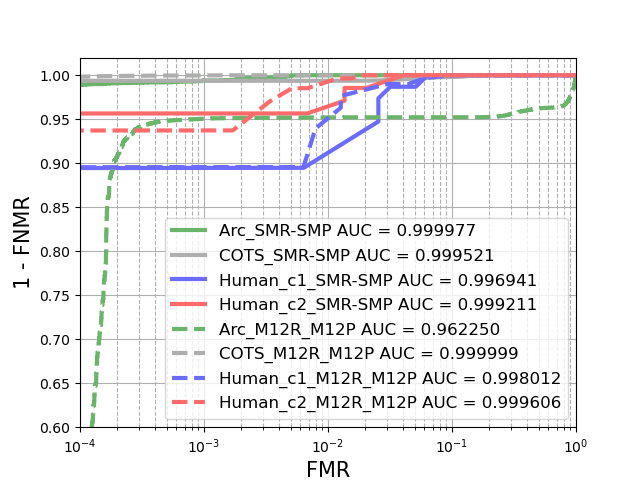}
      \caption{ROC curves of the ArcFace, COTS, Human-c1, Human-c2 on the SMR-SMP and M12R-M12P configurations. While ArcFace loose performance, COTS slightly gain performance when the comparison is between two masked-faces with real masks in comparison to those with synthetic masks.}
      \label{fig:roc_simMaskedMasked_vs_realMaskedMasked}
\end{figure*}

\begin{figure*}[ht!]
     \centering
     \begin{subfigure}[b]{0.32\textwidth}
         \centering
         \includegraphics[width=\textwidth]{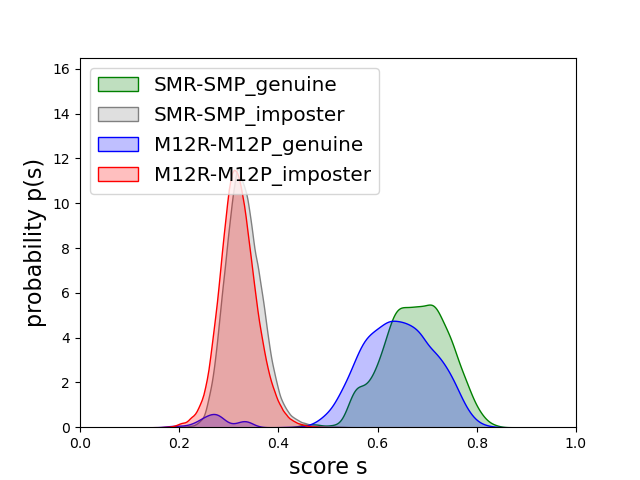}
         \caption{Arc}
         \label{fig:samp:arc222}
     \end{subfigure}
     \hfill
     \begin{subfigure}[b]{0.32\textwidth}
         \centering
         \includegraphics[width=\textwidth]{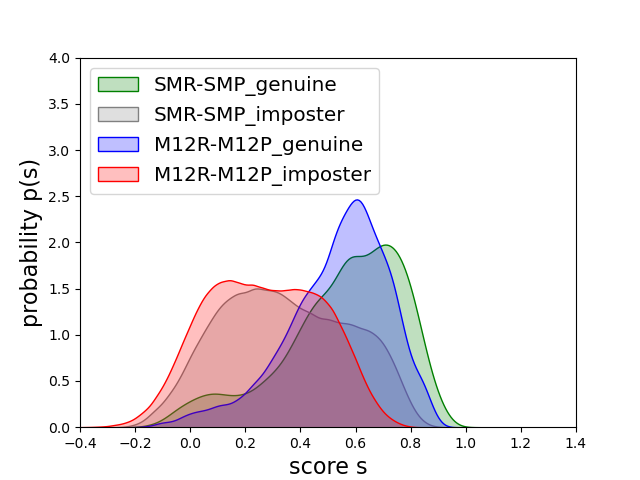}
         \caption{Sphere}
         \label{fig:sam:sphere222}
     \end{subfigure}
     \hfill
     \begin{subfigure}[b]{0.32\textwidth}
         \centering
         \includegraphics[width=\textwidth]{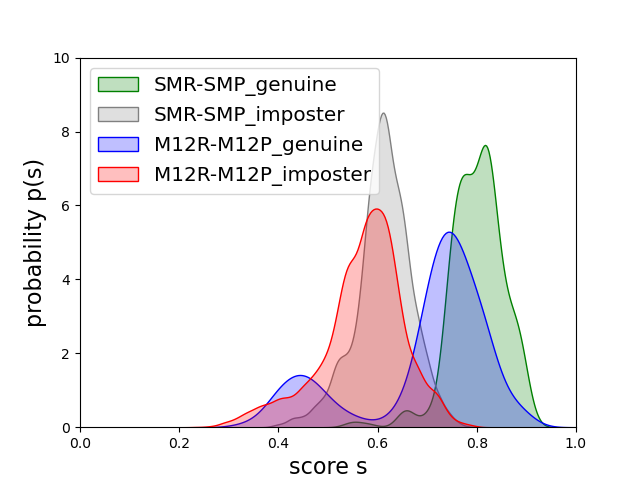}
         \caption{VGG}
         \label{fig:sam:vgg222}
     \end{subfigure}
     \hfill
     \begin{subfigure}[b]{0.32\textwidth}
         \centering
         \includegraphics[width=\textwidth]{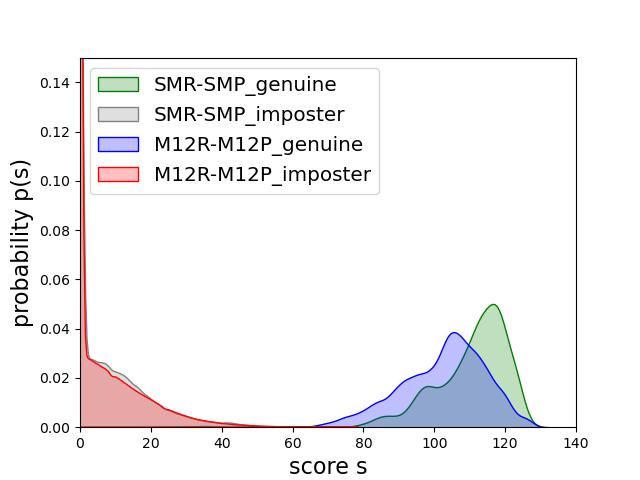}
         \caption{COTS}
         \label{fig:sam:cots222}
    \end{subfigure}    
    \begin{subfigure}[b]{0.32\textwidth}
         \centering
         \includegraphics[width=\textwidth]{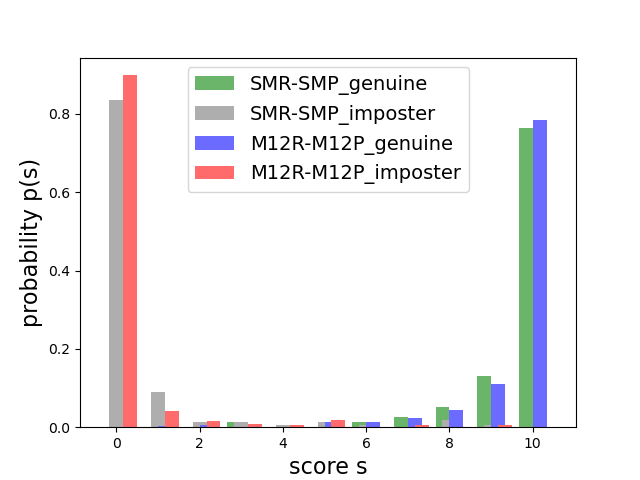}
         \caption{Human-c1}
         \label{fig:sam:hc1222}
     \end{subfigure}
     \hfill
     \begin{subfigure}[b]{0.32\textwidth}
         \centering
         \includegraphics[width=\textwidth]{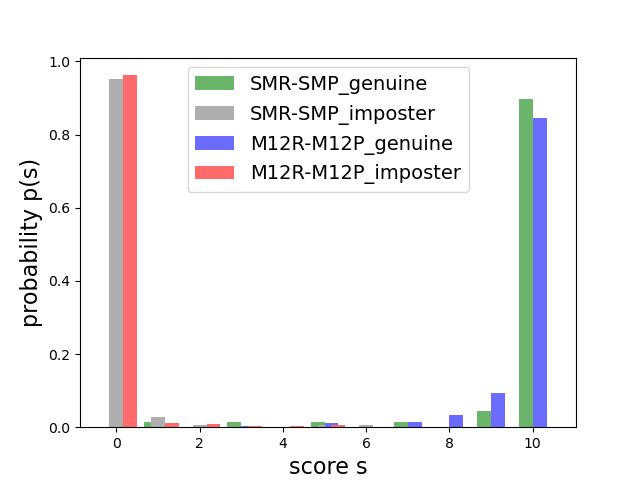}
         \caption{Human-c2}
         \label{fig:sam:hc2222}
     \end{subfigure}
     \caption{The comparison score (similarity) distributions comparing the synthetically masked SMR-SMP genuine and imposter distributions to those of the distributions including real masked M12R-M12P. The real masks induce slightly larger genuine and imposter scores shifts towards the meddle of the distributions, pointing out lower separability than the comparisons of synthetically masked-faces).}
    \label{fig:dist_smrsmp_m12rm12p}
     \hfill
\end{figure*}

\noindent\textbf{How well can a probe with a simulated mask represent the real mask, in the scenario of comparing two masked-faces, for the performance evaluation of human experts and software solutions?}

The verification performances achieved when a pair of synthetically masked-faces are compared (SMR-SMP) are presented in Table \ref{eer_SMR_SMP}. 
In comparison to the performances achieved when the compared pair are wearing real masks (M12R-M12P) in Table \ref{eer_M12R-M12P}, the performances of the low performing automatic solutions (ArcFace, SphereFace, and VGGFace) are less affected by the synthetically masked-faces, i.e. perform better on the synthetically masked pairs. The human experts and the top-performing automatic system (COTS) performances on the other hand were lower when dealing with synthetically masked pairs. A similar trend can be seen in the FDR separability measure. The ROC curves produced by the SMR-SMP and M12R-M12P configurations by the top-performing automatic solutions (ArcFace and COTS) and the two human expert setups are presented in Figure \ref{fig:roc_simMaskedMasked_vs_realMaskedMasked}. There, the human experts' decisions on the synthetically masked pairs, in comparison to the real masked ones, are more accurate at some operational ranges and less or equally accurate at other ranges. The Arcface, on the other hand, performs better on simulated masks.

The genuine and imposter scores distributions of all verification solutions on the SMR-SMP and M12R-M12P are presented in Figure \ref{fig:dist_smrsmp_m12rm12p}. For all automatic solutions, one can notice a shift, in both the genuine and imposter scores of the M12R-M12P comparisons, towards a lower similarity range, in comparison to the SMR-SMP comparisons. This trend is less apparent in the human expert results in the same figure.

Based on the presented results, we can point out that there are differences between the effects of comparing synthetically masked pairs and comparing real masked pairs. This difference is more consistent when it comes to the lower-performing automatic FR solutions. In our experiment, the performance on the M12R-M12P in comparison to that on the SMR-SMP was more accurate when it comes to the top-performing system (COTS) and the human experts.

%

\subsection{Observations of the human experts}
\label{ssec:HEOb}
The human experts performing the experiments were asked to give any observations they might have on the different combinations of the verification process, as well as any general comments. The following observations were made by 4 or more (out of 12) of the human experts and were not contradicted by other experts comments:

\begin{itemize}
    \item Comparisons involving masked-faces were more challenging (less confident decisions) than unmasked pairs comparisons.
    \item The decisions were easier to be made (more confident) when the face crop was wider (crop-2).
    \item Verifying a pair of masked-faces was slightly less challenging than verifying unmasked-to-masked pairs.
    \item Verifying a pair that includes an unmasked reference and a simulated mask probe was slightly less challenging than verifying pairs with real masks probe.
    \item When comparing two masked-faces or unmasked-to-masked faces, making imposter decisions was more confident than making genuine decisions.
    \item The differences in gender and age made it easier to make an imposter verification decision even when masks are worn.
\end{itemize}

The observers' comments 1-4 are directly related to the presented quantitative verification evaluation results and they show a strong correlation to them. Comment 5 correlates to the fact that masked-faces induce larger shifts in the genuine scores towards the imposter range, rather than the imposter scores towards the genuine range.

\subsection{Take-home messages from the experimental results}
\label{ssec:thm}
Based on the experiments conducted in this work, the following main take-home messages can be made:

\begin{itemize}
    \item The verification performance of human experts is affected, in a similar manner to that of automatic FR solutions, by comparing masked probes and unmasked references or comparing pairs of masked-faces. That was demonstrated by the similar trend in performance and separability changes, in most experimental configurations, between human experts and the top-performing automatic FR solutions.
    \item The verification performance of the human experts consistently improves with a wider visible crop that includes areas like the hairstyle, especially when dealing with masked-faces. However, this might not be realistic, as these areas are not stable in applications where the time gap between the references and probes are captured/seen with a large time gap, e.g. border control.
    \item Most of the effect of masked probes (compared to unmasked references) on the verification performance of human experts, such as with automatic FR solutions, is related to relatively large shifts in the genuine scores towards the imposter range. 
    \item The different settings including synthetically created masks show that they are not suitable to represent real masks in the evaluation of human or automatic FR verification performance. This can be related to the limited variation in the synthetic mask design, however, it would be a challenging task (if not impossible) to find the correct synthetic mask variation design. Real masks might also cause variation related to reflection and shadowing.
    \item Comparing pairs of masked-faces in comparison to unmasked-to-masked faces has produced mixed relative verification performances, with the best performing COTS (unlike lower performing systems) achieving higher results and separability on masked pairs and the human experts achieving higher separability on masked pairs.
    \item While comparing unmasked-to-masked faces mainly affected the genuine scores with the imposter scores less affected, comparing pairs of masked-faces have caused shifts in both the genuine and imposter scores towards each other.
    \item Having a masked reference to compare with masked probes is not consistently beneficial. However, having a synthetically masked reference, in this case, have more consistent signs of enhancing the verification results. This might be due to the constant appearance of the simulated masks. While similar real masks might affect the confidence in an imposter decision and dissimilar real masks might affect the confidence of a genuine decision, simulated masks might minimize this problem, especially in the case of imposter pairs.
    \item There is a clear correlation between the main comments made by the human experts on the verification process and the statistical evaluation observations made on their verification performance.
\end{itemize}

These conclusions raise a question regarding the possibility of enhancing the masked face verification performance of human experts through explicit training. They also provide important clues for the development of FR solutions that are robust to masked faces. 
Such solutions recently focused on training FR models that can process both masked and unmasked faces \cite{DBLP:conf/icb/BoutrosDKRKRKFZ21,DBLP:conf/fgr/HuberBKD21} or on reducing the effect of the mask on the face embedding by learning to transfer it into an embedding that behaves similarly to that of an unmasked face \cite{DBLP:journals/corr/abs-2103-01716}.

\section{Conclusion}

This work presented an extensive joint evaluation and in-depth analyses of the face verification performance of human experts in comparison to state-of-the-art automatic FR solutions.
The main motivation behind this effort is the widespread use of face masks as a preventive measure to the COVID-19 pandemic.
We analyzed the correlations between the verification behaviors of human experts and automatic FR solutions under different settings.
These settings involved unmasked pairs, masked probes and unmasked references, and masked pairs, with real and synthetic masks.
We formed a set of take-home messages that addressed the relative effect of masked-faces on the human verification behavior, the use of synthetic masks in evaluation efforts, and the sanity of having a masked reference, among others. We additionally list a set of human experts' observations on the verification process.

\section{Acknowledgments}\label{sec11}

This research work has been funded by the German Federal Ministry of Education and Research and the Hessen State Ministry for Higher Education, Research and the Arts within their joint support of the National Research Center for Applied Cybersecurity ATHENE.

{\small
\bibliographystyle{IEEEtran}
\bibliography{main}
}

\end{document}